\documentclass[generic]{imsart}
\arxiv{1510.00757} 

\usepackage{float,lscape}

\RequirePackage[OT1]{fontenc}
\RequirePackage{amsthm,amsmath,amssymb,amsfonts}
\RequirePackage[numbers]{natbib}
\RequirePackage[colorlinks,citecolor=blue,urlcolor=blue]{hyperref}
\RequirePackage{mathtools,bbm}
\RequirePackage{placeins}

\usepackage{footnote}
\usepackage{rotating,booktabs}

\DeclarePairedDelimiter{\ceil}{\lceil}{\rceil}

\DeclareMathOperator*{\argmax}{arg\,max}

\startlocaldefs
\numberwithin{equation}{section}
\theoremstyle{plain}

\endlocaldefs

\begin{document}

\begin{frontmatter}
\title{A Survey of Online Experiment Design with the Stochastic Multi-Armed Bandit}
\runtitle{Survey of Multi-Armed Bandit Experiments}

\begin{aug}
\author{\fnms{Giuseppe} \snm{Burtini}\thanksref{}\ead[label=e1]{g.burtini@alumni.ubc.ca}},
\author{\fnms{Jason} \snm{Loeppky}\thanksref{T1}\ead[label=e2]{jason.loeppky@ubc.ca}}
\and
\author{\fnms{Ramon} \snm{Lawrence}\thanksref{T1}\ead[label=e3]{ramon.lawrence@ubc.ca}}

\address{Departments of Computer Science and Statistics,\\ 
University of British Columbia - Okanagan, \\
Kelowna, Canada\\
\printead{e1,e2,e3}}

\thankstext{T1}{Loeppky and Lawrence were partly supported by Natural Sciences and Engineering Research Council of Canada Discovery Grants, grant numbers {RGPIN-2015-03895} and {RGPIN-341202-12} respectively.}
\runauthor{G. Burtini et al.}

\affiliation{University of British Columbia}

\end{aug}

\setcounter{tocdepth}{4}

\begin{abstract}
Adaptive and sequential experiment design is a well-studied area in numerous domains. We survey and synthesize the work of the online statistical learning paradigm referred to as \emph{multi-armed bandits} integrating the existing research as a resource for a certain class of online experiments. We first explore the traditional stochastic model of a multi-armed bandit, then explore a taxonomic scheme of complications to that model, for each complication relating it to a specific requirement or consideration of the experiment design context. Finally, at the end of the paper, we present a table of known bounds of regret for all studied algorithms providing both perspectives for future theoretical work and a decision-making tool for practitioners looking for theoretical guarantees.
\end{abstract}


\tableofcontents
\end{frontmatter}

\section{Introduction}
The real world has a wealth of circumstance where an individual, business or researcher must simultaneously explore their surroundings, options or choices while also maintaining or maximizing some variable: their output, well-being or wealth. Indeed the pursuit of a good life can be represented as finding a balance between ``exploration,'' or the investigation of new options and ``exploitation,'' the utilization of the knowledge one has already accrued. People make these decisions on a regular basis, from career and education (``should I continue to invest in education, exploration, or begin exploiting the knowledge I have accrued?'') to shopping and selection (``should I purchase a product I have already experienced and am satisfied with, or should I explore other options?'') and everything in between. 

These problems have been studied extensively in a human decision making\footnote{Often discussed using the language ``naturalistic decision making'' (NDM)} context, especially with regard to marketplace decisions \cite{Bell:1985,Erdem/Keane:1996,Lipshitz/Strauss:1997}, a game theory ``maximizing'' context \cite{Osborne/Rubinstein:1994}, and the positive psychology of increasing happiness through decision making \cite{Hastie/Dawes:2010,Kahneman:2003,Kahneman/Klein:2009}. This type of decision making also arises in learning problems of the statistical and machine learning variety and is represented within a broad class of problems called \emph{reinforcement learning.} One common form of this \emph{exploration vs. exploitation tradeoff} is called the \emph{multi-armed bandit problem} where the world can be imagined as a collection of slot machines, each with a different but unknown payoff, and the player must play, repeatedly, to maximize his wealth. This is the form of the \emph{sequential decision making problem} for exploration vs. exploitation that we will explore in this work. 

These sequential decision making problems appear foremost in the machine learning and artificial intelligence literatures. A strong conceptual framework has been developed, algorithms discovered and analyzed and tools produced and compared within this literature since as early as 1933 \cite{Thompson:1933}. This survey aims to integrate the advances in the area of multi-armed bandits and sequential decision making with the literature of experiment design, especially sequential experiment design, providing a statistical perspective on the state-of-the-art and presenting a unified discussion of response-adaptive experiments with multi-armed bandits. 

\subsection{Motivating Applications}
An exemplary motivating application is the area of medical trials. Pragmatically, in our representation of a medical trial, the goal is to conduct an experiment to measure, to a sufficient confidence, the effect size of a particular medication compared to a placebo or control group. Secondary goals such as measuring worst-case performance and both objective and subjective side-effects of the medication are also important. Observations in medical trials are expensive, both because of the research nature of the pharmaceuticals often under experimentation and because of the necessity to recruit qualified patients, in many cases, constrained to be within a single geographical area where sample sizes on the order of tens may be all that is available.

In a traditional experiment design, \emph{efficiency} relates the number of observations (patients) necessary to achieve a given level of confidence on any estimators involved. This definition of efficiency often means a medical trial (for a single drug) will operate by allocating, generally randomly or pseudo-randomly, one half of the patients to a placebo group and one half of the patients to a treatment group, then computing traditional statistics on the aggregate results.

Sequential experiment design and adaptive designs, at first two-stage approaches, have been considered in the medical literature since as early as 1960 \cite{Armitage:1960, Anscombe:1963, Colton:1963, Day:1969}. A recurring ethical confound arises when discussing clinical experiment design which motivates the sequential design: estimator efficiency is at odds with saving patients in the short-term. During the experiment, information can be revealed to the experimenter which provides evidence about the uncertainty and expected value of each estimator, information which if available \emph{ex-ante}, may have changed the design of the experiment. A utility or outcome-weighted approach to experiment design would certainly design a different experiment in the event revealed information increased the pre-experiment knowledge in favor of one medication over another.

Techniques alluding to similar considerations as the multi-armed bandit problem such as the \emph{play-the-winner} strategy \cite{Tolusso:2012} are found in the medical trials literature in the late 1970s \cite{Wei/Durham:1978,Rosenberger:1999}. In the 1980s and 1990s, early work on the multi-armed bandit was presented in the context of the \emph{sequential design of experiments} \citep{Robbins:1952,Lai/Robbins:1985} and \emph{adaptive experiment design} \cite{Berry/Fristedt:1985}.  

Adaptive design research has continued independently, especially in the context of clinical trials. Modern adaptive and sequential designs include drop-the-losers designs, where certain treatments are dropped or added in response to their response data; group-sequential or stagewise designs which act similar to the epoch-designs discussed in multi-armed bandits; and sample-size reestimation techniques which allow sample sizes to be adjusted throughout the research process. Thorough overviews have been recently given by \cite{Chow/Chang:2008} in 2008 and \cite{Sverdlov/Wong/Ryeznik:2014} in 2014. 

Two important subclasses of adaptive design exist which map intuitively to subproblems of the multi-armed bandit. This includes the adaptive dose finding design often used in early clinical trials to identify minimum and maximum dosages for pharmaceuticals, and the biomarker-adaptive design which is used in a number of ways to improve research outcomes with respect to observable covariates in the patient. Adaptive dose finding designs can be seen as a special case of continuum-armed bandits and biomarker-adaptive designs can be seen as a special case of contextual bandits, where biomarkers are observed as a possible contextual variable in determining the response variable.


Since the early bandits research, a policy or decision-theoretic lens has often been used to explore multi-armed bandits, characterizing the problem as a recurrent policy selection to maximize some utility function or minimize some regret function and becoming somewhat disconnected from the experiment design literature. In the remainder of this work, we integrate the rapid advances in {multi-armed bandits} with a perspective of online experiment design. Recently, a number of works in multi-armed bandits have shown that even when the objective function is specified solely as the maximization of some outcome variable, reliable experiment results can be found to a sufficiently high degree of confidence \cite{Kuleshov/Precup:2014,Kaptein:2015}.

As noted by \cite{Kuleshov/Precup:2014}, existing adaptive designs often offer limited theoretical guarantees on outcomes and patient welfare, while multi-armed bandits work has numerous theoretical guarantees, many of which are fundamental to the ethical and practical considerations in clinical and other experiment design work. Furthermore, many areas of design research such as the biomarker-adaptive designs produce partial solutions to a contextual problem that is more completely explored in the multi-armed bandits literature. 


\subsection{The Multi-Armed Bandit Model}

Chronologically, \cite*{Robbins:1952} introduces the idea of the important tradeoff between exploration and exploitation in recurring decision problems with uncertainty, building on the prior \emph{sequential decision problem} work of \cite{Wald:1947} and \cite{Arrow:1949}. \cite*{Lai/Robbins:1985} produce the first asymptotic analysis of the objective function of such a decision process showing a bound of $O(\log t)$ for the regret of the standard stochastic, finite-armed, multi-armed bandit, and produced a regret-efficient solution where the rewards of a given arm are stationary and \emph{independent and identically distributed} (i.i.d.) with no \emph{contextual information} or covariates. In the coming sections, we will explore some of the large number of algorithms which have been proposed since \cite{Lai/Robbins:1985} including  $\epsilon$-exploration approaches \citep{Watkins:1989,Vermorel/Mohri:2005}, upper-confidence bound techniques \citep{Auer/Cesa-Bianchi/Freund/Schapire:2002}, probability matching techniques \citep{Agrawal/Goyal:2012} and others. Many variants of the initial problem have also been investigated in the literature including the many- or infinite-armed, adversarial, contextual, and non-stationary cases.

The terminology ``bandit'' originates from the colloquialism ``one-armed bandit'' used to describe slot machines. It is important to note immediately that this stylized terminology suggests the negative expected value embodied in the context of the word ``bandit,'' despite that not being a requirement, nor being common in most applications of the model. In one common formalization of the multi-armed bandit model, the player can sequentially select to play any arm from the $K\geq1$ arms and obtain some payoff $x\in\mathbb{R}$ with some probability $p\in[0,1]$. This specific formalization has explicitly defined binomial arms (fixed payoff, random variable payoff probability), however in our interpretation of the problem, the distribution on the arms can (and often is) be from any distribution. 

In all variants of the bandit problem only the payoff for the \textbf{selected} arm at any time step is observed and \emph{not} the payoffs for non-selected arms. This is the ``partial information'' nature of the bandit problem which distinguishes the problem from generalized reinforcement learning, where ``full information'' (observing the payoff of all arms, whether selected or not) is usually assumed. This is precisely the property which makes the multi-armed bandit an appropriate tool for experiment design, a treatment is selected and observed and the results of that treatment are provided as feedback to the algorithm.

In many of our considered variants of the problem, the reward distribution itself is not assumed, however, rewards are often assumed to be \emph{i.i.d.} across arms and across prior plays. Some variants of the problem relax both of these assumptions. Similarly, it is generally assumed that both $K$, the number of machines (or ``arms'') and $x$, the payoff function are finite and stationary (time-invariant), however, variants relaxing these assumptions have also been considered. An important variant of the problem introduces the concept of a known, finite time-horizon, $H$ to bound play time. Two important variants, the contextual problem and the adversarial problem, remove assumptions such that there are fewer or even no statistical assumptions made about the reward generating process.

To reiterate the nature of the problem we are considering, we provide the following summary formalization. A (finite-armed, stochastic) multi-armed bandit problem is a process where an agent (or forecaster) must choose repeatedly between $K$ independent and unknown \emph{reward distributions} (called \emph{arms}) over a (known or unknown) time horizon $H$ in order to maximize his total reward (or, minimize some total \emph{regret}, compared to an oracle strategy which knows the highest reward answer at any point in time). At each time step, $t$, the \emph{strategy} or \emph{policy} selects (\emph{plays}) a single arm $S_t$ and receives a reward of $x_{S_t,t}$ drawn from the $i$th arm distribution which the policy uses to inform further decisions. 


\section{Considerations of Multi-Armed Bandits}

Most sequential experiment designs can be thought of as multi-armed bandits. The primary difference is motivated by whether the goal is to identify the best of a set of variants according to some criteria (i.e., \emph{best-arm identification}) or to conduct multiple post-hoc tests of statistical significance (e.g., each arm versus each other arm). In algorithms designed for the multi-armed bandit, exploring suboptimal arms according to our objective criteria is non-desirable. This concept referred to as \emph{regret}, makes the multi-armed bandit less useful for analyses which involve comparing arm-to-arm than analyses which aim for best-arm identification, however, \cite{Kaptein:2015} shows a real-world example demonstrating that arm-to-arm comparisons are still plausible in a multi-armed bandit environment while maximizing within-experiment utility.

\subsection{Measures of Regret}
There are a number of possible objective functions to evaluate a bandit process in an empirical, non-adversarial setting that have been used within the literature. In general, regret-based objectives are loss functions which compare the performance of a process with an oracle over a finite and discrete time horizon $t=1,...,H$. The definitions vary on how the stochastic nature of the process is considered and what information and selections are made available to the oracle. The sequence of arms selected $S_t$ can itself be a random variable (such as in the Thompson sampling case discussed later). 

For most applications of multi-armed bandit models this is the most important conceptual variable to consider, but in the case of the experiment design literature it may not be: estimator variance minimization and the ability to accurately capture measures such as effect sizes can be more important. Nonetheless, it is unfortunate that regret is not consistently defined in the literature as a single variable of interest, but generally takes one of a variety of forms. In a pragmatic sense, regret has the same meaning as it has in English: ``the remorse (losses) felt after the fact as a result of dissatisfaction with the agent's (prior) choices.'' We summarize and formalize some of the varieties of regret which both arise in the literature and are relevant to the problem of experiment design with multi-armed bandits. For a comprehensive discussion of the differing measures of regret, see \cite{Burtini/Loeppky/Lawrence:2015b}.

\begin{enumerate}
\item \textbf{Expected-expected regret}. This is the difference in the payoff of the selected arm, in expectation over the distribution of that arm, and the payoff of the optimal arm as given by a prescient oracle, in expectation over the distribution of that arm. Importantly, this measure includes no unnecessary variation, but cannot be computed without a complete understanding of the definition of all arms.

\begin{equation}\bar{R}^{E} = \sum_{t=1}^H \left( \max_{i=1,2,...,K} \mathrm{E}[x_{i,t}]\right) - \sum_{t=1}^H \mathrm{E}[x_{{S_t},t}]\end{equation}

\item \textbf{Expected-payoff regret}.  This is the same measure as expected-expected regret, but without an expectation on the right hand term. That is, the actually received reward is used rather than the expectation of the selected arm. This is important in environments where measuring the variation between runs is meaningful, especially in risk-aware cases where maximal drawdown or maximal loss may be motivating factors.

\begin{equation}
	\bar{R}^{P} = \sum_{t=1}^H \left( \max_{i=1,2,...,K} \mathrm{E}[x_{i,t}] \right) - \sum_{t=1}^H x_{{S_t},t}.
\end{equation}

\item \textbf{Suboptimal plays}. This is a more egalitarian perspective on regret for many applications, where the measure is not proportional to the distance between distributions. In the two-armed case, this is simply a matter of scale, but in larger problems this can mean that two very close means are weighted equally compared to two distant means as both being \emph{suboptimal}.

\begin{equation}
N_{\vee} = \sum_{t=1}^H \mathbbm{1}[S_t \neq \argmax_{i=1,2,...,K} \mathrm{E}[x_{i,t}]].
\end{equation}

\end{enumerate}

In general, regret measures can only be computed in a simulation environment because of their requirement of an oracle which can unambiguously identify the optimal arm (and in the case of the expectation measures, to also take expectations over those distributions). Burtini et al. \cite{Burtini/Loeppky/Lawrence:2015b} introduces a measure called \emph{statistical regret} which removes the oracle requirement:

\begin{enumerate}
  \setcounter{enumi}{3}

\item \textbf{Statistical regret}. Statistical regret uses the idea of a confidence or predictive bound at time $t$ to replace the oracle and evaluate the plays at times $j = 0, ..., t$. This utilizes the \emph{best information available} to judge actions the policy chose prior to that information being available. To compute parametric statistical regret, we need an interval $(L_{t,\gamma}(\tilde{X}_{i}), U_{t,\gamma}(\tilde{X}_{i}))$ such that,

\begin{equation*}
\mbox{Pr}(L_{t,\gamma}(\tilde{X}_{i}) < \tilde{X}_{i} < U_{t,\gamma}(\tilde{X}_{i})) = \gamma
\end{equation*}

If such an interval cannot be computed due to limitations of the parametric modelling process, a bootstrap \citep{Efron:1979} approach can be substituted.

\begin{equation}
\tilde{R}^P_\gamma = \left(\sum_{j = 0}^t (\max_{i} L_{t,\gamma}(\tilde{X}_{i}) - x_{S_j}), \sum_{j = 0}^t (\max_{i } U_{t,\gamma}(\tilde{X}_{i}) - x_{S_j})\right)
\end{equation}
\end{enumerate}

\subsection{Variance and Bounds of Regret}

Similarly to the well-known bias/variance tradeoffs in the estimator selection literature, variance minimization is an important parameter for many applications. In the most extreme case, a high-variance algorithm will be  unacceptable in a practical sense even if it had a minimal expectation regret. Certain choices in the measurement of regret are more appropriate if one chooses to measure variance than others, for instance, selecting expected-payoff regret allows one to measure the variance of an individual payoff (e.g., a next play payoff variance) while expected-expected regret will only allow measurement of variance across repeated plays of the full sequence.

Often, rather than measuring variance of a policy, a \emph{high-probability bound} on the regret is proposed. High-probability bounds arise from the ``probably approximately correct'' learning (PAC learning) literature \cite{Valiant:1984} which provides a framework for evaluating learning algorithms with traditional asymptotic measures from computational complexity theory. The language $\mbox{PAC}_{\epsilon,\delta}$-learnable provides that an algorithm exists such that it can learn the true value with an error rate less than $\epsilon$ with probability of at least $1-\delta$ in the limit or within finite number of fixed iterates.

High-probability bounds introduce this PAC learning framework to regret analysis. The primary difference from the regret {in expectation} is that a high-probability bound considers the potential variation in return, which can be (and is, in many cases) large, a factor which is very important in medical and many financial contexts, where stakes are high. Intuitively, a high-probability bound provides a function that can be used to evaluate a bound at a given probability. This is a statement of the form $P[R(n) > b_{\textrm{\footnotesize high\_probability}}] \leq \delta$ where $R(n)$ is the regret after $n$ plays, $b_{\textrm{\footnotesize high\_probability}}$ is the high-probability bound and $\delta$ is the probability with which we evaluate the ``high'' probability. The parameter $\delta$ is often picked to be a function of the form $n^{-c}$ for a fixed constant $c$ \cite{Aspnes:2014}.

To provide an intuitive understanding of high-probability bounds compared to expectation regret, consider the slot-playing $\epsilon$-first example: imagine we have two slot machines to pick between, and we explore 10 times (5 each) to measure our empirical estimates of each arm then exploit the best measured machine forever. In expectation the best machine will be picked, however, randomness and the small number of plays may result in a single win dominating the results and causing the estimate of the best machine to be incorrect. While the expectation of regret in this model may be low, the variance of regret is high: in many plays, the regret will tend to infinity.  The high-probability bound in general, will say that for $p$\% or fewer of the repeated plays of the game, the regret will exceed $b$. In this specific example, it is likely the bound will be very weak for any reasonable $\delta$, as in this particular strategy, a small number of lucky early explorations will result in suboptimal plays forever, accruing large regret.

High-probability bounds are asymptotic. A related form of bound for multi-armed bandits are the finite time bounds. These are arguably more important for applications. A finite time bound is a function of $t$, the number of trials so far, which provides a probabilistic or concrete limit on the regret at that time $t$. These are often special forms of high-probability bound themselves, holding in high-probability but non-asymptotic cases.




\subsection{Stationarity of the Problem}

One of the strongest assumptions of many statistical models, including most variants of the multi-armed bandit problem, is that the underlying distributions and parameters are \emph{stationary}. In many contexts, including the context studied here, this is not a reasonable assumption: the state of the world is changing around the learning process and in our context, the best arm in one time period may not be the best arm in another. Non-stationary problems are in general challenging for algorithms that make stationarity assumptions, whether explicit or implicit, as the real world performance of any such policy can continuously degrade in response to unconsidered changes in the distribution. In particular, rapid changes of the distribution and switching-type models (day, night; seasonal; or any other repeatedly changing, but unmodelled, confounding factor) have extremely poor performance on many fixed policies.

Some variants of the model, known generally as non-stationary bandit models have been proposed with drifting or step-function changing parameters.  A simple solution to deal with non-stationarity is to allow the data to ``decay'' out of the model with a time-weighted component, however, this solution requires an accurate model of the appropriate rate of decay to be efficient. We discuss algorithms for the non-stationary variant of the policy in a later section, especially in the context of running an experiment in an ongoing indefinite fashion. 

\subsection{Feedback Delay}

Feedback delay is an under-considered issue in the multi-armed bandit literature, especially with regard to extremely long feedback delays. In the simple multi-armed bandit model, an arm is selected, played and reward received before the next iterate takes place. In many real world applications this is not the case - an arm must be selected, but the reward may not be received until much later, by which time many other iterates may have taken place. This is an important consideration in the design of a clinical trial, for instance, as patients may come in batches and treatment may take on the order of months before the response is fully observed. 

Little research exists on the topic of quantifying and understanding feedback delay in most multi-armed bandit algorithms, however \cite{Kuleshov/Precup:2014} have shown that clinical trials, especially late stage comparative trials can be conducted successfully under the case of feedback delay with the multi-armed bandit. Importantly, a variant of multi-armed bandits \emph{with multiple plays}, which we explore later in this paper can be used to model certain classes of feedback delay when the delay is known.

\section{Algorithms for the Application of Multi-Armed Bandits}

The most studied variant of the model is the traditional model described earlier, with a discrete, finite number of arms ($K$) for the agent to choose between at each time step $t$. There are a large number of techniques for solving this variant.

A strategy or algorithm used to solve the multi-armed bandit problem is often called a \emph{policy}. First, we explore a theoretically popular Bayes-optimal policy called the Gittins index, introduced in 1979. Due to the computational limitations of Gittins indicies, we then explore the evolution of the literature hence, discussing a simple (in implementation) set of policies, the $\epsilon$ policies, dependent on a parameter $\epsilon$ which determines how much exploration takes place, a set of policies called UCB-strategies (upper confidence bound strategies), based on an observation by \cite{Agrawal:1995} and later \cite*{Auer/Cesa-Bianchi/Fischer:2002} on utilizing {upper confidence bounds} which efficiently approximate the Gittins index, a variety of standalone policies and finally probability matching policies which rely on the idea of matching the probability of success with the probability of drawing that arm. Strategies like $\epsilon$-based strategies that maintain an ongoing distinction between exploitation and exploration phases are called semi-uniform. 

\subsection{Gittins index}

An early solution to the multi-armed bandit problem requires the experimenter to treat the problem as a Markov decision process and use work similar to \cite{Puterman:2014} to solve it. Unfortunately, computational considerations and the \emph{curse of dimensionality} \cite{Chakravorty/Mahajan:2013} produce a problem which is exponential in the number of possible realizations and arms of the bandit process. To avoid this complexity, \cite{Gittins/Jones:1979} and \cite{Gittins:1979} show that the problem can be reduced to solving $n$ 1-dimensional problems, computing the so-called \emph{Gittins index} and selecting the arm with the highest index value. The index is given as,

\begin{equation}
\nu^i(x^i)  = \sup_{\tau > 0} \frac{ E\left[ \sum_{t=0}^\tau  \beta^t r^i(X^i_t) | X^i_0 = x^i \right] }{ E\left[ \sum_{t=0}^\tau \beta^t | X^i_0 = x^i \right]} 
\end{equation}

Where $x^i$ is the \emph{state} of the arm $i$, $\tau$ is a positive integer referred to as the \emph{stopping time} and $\nu^i$ is the value of the Gittins index for arm i. The stopping time $\tau$ represents the first time at which the index for this arm may no longer be optimal, that is, the index is no greater than its initial value. The decision rule is then a simple $i^t = \argmax_i \nu^i(x^i)$ to be computed at each time. Fundamentally, the numerator represents the expected total $\beta$-discounted reward over the next $\tau$ plays and the denominator is a $\beta$-discount with respect to time, producing an estimated average discounted reward per play $\nu^i$. Simply, it can be shown \cite{Gittins/Jones:1979} that it is optimal to play the arm with the current highest index.

The motivating perspective of the Gittins policy is different than our perspective applied to other algorithms for the multi-armed bandit. Specifically, when discussing the Gittins index, we treat the arms as Markov decision processes (often which are evolving in a \emph{restless} or \emph{nonstationary} sense) rather than statistical distributions. This perspective becomes limiting when we consider bandit problems where the state-space may be infinite, such as in the case the arms take on a Bernoulli process. 


While the Gittins policy is efficient (e.g., \cite{Weber:1992, Tsitsiklis:1994, Whittle:1980, Frostig/Weiss:2013}) in terms of using the available knowledge to minimize regret, it has a set of serious limitations: the computational complexity of maintaining the indices is high \cite{Katehakis/Veinott:1987} and is practically constrained to a limited set of known distributions representable as a finite state-space model. These constraints lead to the work of \cite{Katehakis/Veinott:1987,Brezzi/Lai:2002} showing how the Gittins policy can be efficiently approximated by techniques of optimism like UCB.  In general, we favor the statistical approach to the multi-armed bandit in the rest of this work, but the motivating example of the Gittins index has remained an important part of the development of bandit models to date.

\subsection{$\epsilon$-greedy}

The $\epsilon$-greedy approach appears to be the most widely used simple strategy to solve the simple stochastic, i.i.d. form of the (discrete) multi-armed bandit model in practice. The strategy, in which the agent selects a random arm $0\leq\epsilon\leq1$ fraction of the time, and the arm with the best observed mean so far otherwise, was first presented by \cite{Watkins:1989} as a solution to the equivalent one-state Markov decision process problem\footnote{Watkins' motivation was in modelling learning processes in the real world, not for machine learning. The distinction does not appear to be important for the particular policy he devises.}. The choice of $\epsilon$ and strategy for estimating the mean is left to the application.

$\epsilon$-based strategies have been well studied.  \cite{Even-dar/Mannor/Mansour:2006} show that after $O\left(\frac{K}{\alpha^2} \log \frac{K}{\delta}\right)$ random plays an $\alpha$-optimal arm will be found with probability greater than $1-\delta$, a result that applies to all major $\epsilon$ strategies.


\subsubsection{Constant $\epsilon$}
With a constant value of $\epsilon$, a linear bound on regret can be achieved. Constant $\epsilon$-greedy policies are necessarily suboptimal, as a constant $\epsilon$ prevents the strategy, in general, from asymptotically reaching the optimal arm \cite{Vermorel/Mohri:2005}. That is, even after strong knowledge is acquired, the strategy will continue to behave randomly some $\epsilon$ fraction of the time.

\subsubsection{Adaptive and $\epsilon$-Decreasing}
One of the more salient variants of $\epsilon$-greedy is the $\epsilon$-decreasing strategy. In a stationary, finite horizon environment, it is logical to have a policy do more exploration early and more exploitation as it becomes more confident about its knowledge or as it gets closer to its horizon. This can be implemented with a variance weighted strategy or by simply decreasing $\epsilon$ according to some rule (time, observations, etc.). In known-horizon environments, $\epsilon$-decreasing policies can weight the rate exploration as a function of the remaining horizon available, though no known work has explicitly defined the correct functional form to do so.

A simple $\epsilon$-decreasing strategy is natural and given by \cite{Vermorel/Mohri:2005} which defines $\epsilon(t)$ as the value of $\epsilon$ after $t$ plays as $\min(1, \frac{\epsilon_0}{t})$ where $\epsilon_0$ is left as a choice to the user. A similar strategy is called GreedyMix and analyzed in \cite*{Cesa-Bianchi/Fischer:1998} where $\epsilon(t)$ (referred to as $\gamma$) is defined as $\min(1, \frac{5K}{d^2}\cdot\frac{\ln(t-1)}{t-1})$ where $0<d<1$ is a constant picked by the user\footnote{Note that by letting $\epsilon_0 = \frac{5K}{d^2}$ GreedyMix is similar to the Vermorel and Mohri strategy, but not the same, as the rate of decrease is $\frac{\ln(t-1)}{t-1}$.}. GreedyMix is shown to have regret on the order of $\ln(H)^2$ for $H$ trials for Bernoulli- and normally-distributed bandits. Selection of $d$ is left to the reader, and performance degrades if a sub-optimal value of $d$ is selected.

An interesting result regarding $\epsilon$-decreasing policies is given by \cite*{Auer/Cesa-Bianchi/Fischer:2002} with the simulations on a policy called $\epsilon_n$-greedy. $\epsilon_n$-greedy is the generalized form of $\epsilon$ greedy where the fraction of exploration is a function of the time step. At each time step $t$, we select the $\epsilon_t = \epsilon(t)$. By defining $\epsilon(t) \equiv \min \left\{1, \frac{cK}{d^2t} \right\}$ and correctly selecting an unknown parameter $c>0$ and a lower bound $0<d<1$ on the difference between the reward expectations of the best and second best arms, we get a policy which has an expected regret of $O(\log H)$. Unfortunately, as noted in \cite{Auer/Cesa-Bianchi/Fischer:2002} this result is not of a lot of practical use, for the same reason GreedyMix lacks practicality: the selection of the constant factors $c$ and $d$ are dependent on the underlying distribution which we are trying to estimate and the performance degrades rapidly in the incorrectly tuned case. A theoretical, but not practical, extension of this strategy is one where $\epsilon(t)$ is correctly chosen for each time step to properly account for the current uncertainties; this strategy is guaranteed to converge in an optimal number of trials in expectation. A variant of this strategy is made practical by the algorithm referred to as POKER and explored later in this section.

\subsection{$\epsilon$-first}
In the non-academic web optimization and testing literature, $\epsilon$-first is used extensively, generally for 2-armed bandits and is widely known as ``A/B testing''\footnote{The extensive toolsets available for automating this testing often perform ``A/B testing'' incorrectly. Specifically, they perform testing with repeated testing without an appropriate multiple testing significance adjustment. It is up to the user, who is generally not expected to be familiar with the statistics involved, to behave appropriately to maintain the assumptions of the model. Many researchers have addressed the multiple testing issue, for an overview of the problem see \cite*{Jennison/Turnbull:1990}; for an review of strategies for correcting multiple testing errors, see \cite*{Hsu:1996} or \cite*{Westfall/Young/Wright:1993}. Indeed, the most fragile of these toolsets offer functionality to make decisions ``upon reaching significance'' (using an unadjusted measure of significance) which suggests a significance test after every trial: the worst form of the multiple-testing problem, resulting in a false positive rate which increases as the number of trials increases \cite{Jennison/Turnbull:1990}.}. In $\epsilon$-first, the horizon, $H$, should be known \emph{a priori}\footnote{If it is not known, the \emph{number} rather than fraction of exploratory plays will have to be selected. In an environment where the horizon is unknown, it is unlikely this will be a good algorithm as it will be nearly impossible to estimate how much exploration versus exploitation will produce the best result.}. The first $\epsilon \cdot H$ plays are called the exploration phase, and the agent picks arms uniformly randomly, producing an estimate of each arm's payoff. In the remaining $(1-\epsilon) \cdot H$ plays, called the exploitation phase, the agent strictly picks the best empirically estimated arm. 

An $\epsilon$-first strategy is superior to an $\epsilon$-greedy strategy when the horizon is fixed and stationarity on the arms can be assumed as the estimates produced during exploration are expected to be better for a larger number of plays and thus fewer suboptimal plays will be expected. Compared to $\epsilon$-greedy, $\epsilon$-first is vulnerable to non-stationarity of the reward distribution, because all learning takes place ``upfront'' and the knowledge is available for more iterates. $\epsilon$-first is also vulnerable to errors in estimating the \emph{time horizon}, the number of trials remaining.

\subsubsection{Multiple Epoch}

A variant of the $\epsilon$ algorithms is the multiple epoch approach. Multiple epoch approaches can be applied to many multi-armed bandit policies (e.g., \cite{Langford/Zhang:2007}) but they are largely unstudied in non-$\epsilon$ approaches. They may show promise in non-stationary bandit cases where the epoch length (and data decaying) can be used to control for the maximum deviation. In the multiple epoch approach, we divide our total time horizon (known or unknown, finite or infinite) in to epochs of an integer length. The respective policy is then applied \emph{within} the epoch. For example, in the $\epsilon$-first strategy, this eliminates some of the vulnerability to non-stationarity and horizon-unawareness by allowing learning to take place at spaced periods within the total time.

\subsection{UCB1}
Much of the research in regret bounds demonstrates regret that is logarithmic (``optimal'') only asymptotically. \cite{Auer/Cesa-Bianchi/Fischer:2002} present an algorithm originating in \cite{Agrawal:1995} called UCB1 which achieves expected logarithmic regret uniformly over time, for all reward distributions, with no prior knowledge of the reward distribution required. UCB1 is the first strategy we have discussed that is not a \emph{semi-uniform} strategy, that is, it does not maintain a distinction between an exploratory phase and an exploitation phase, choosing instead to optimize how exploration happens at each individual iterate. UCB1 belongs to the general family of \emph{upper confidence bound} (UCB) algorithms, first proposed in \cite{Lai/Robbins:1985} but developed extensively in \cite{Auer/Cesa-Bianchi/Fischer:2002}. UCB algorithms take the form of picking the arm which maximizes a surrogate function, i.e., they pick,

\begin{equation}
	i = \argmax_i \mu_i + P_i
\end{equation}

where $\mu_i$ is the \emph{average function} which estimates the mean payoff of arm $i$ and $P_i$ is the \emph{padding function} which generally takes the form of an approximation of the uncertainty on $\mu_i$. The primary contribution of variants of the UCB algorithms is the selection of $P_i$. 

For convenience, let $\Delta_i$ be defined the same way as in \cite{Auer/Cesa-Bianchi/Fischer:2002}: $\Delta_i \equiv \mu^* - \mu_i$ where $\mu^*$ represents the mean reward expected from the optimal arm and $\mu_i$ represents the current reward expectation for arm $i$.

UCB1 begins by playing each arm once to create an initial estimate. Then, for each iterate $t$, arm $i$ is selected to achieve the maximum value $\max_i \bar{x}_i + \sqrt{\frac{2 \ln t}{n_i}}$ where $\bar{x}_i$ is the average observed reward of arm $i$ thus far (the empirical mean) and $n_i$ is the number of times arm $i$ has been played. The second term in this equation acts as an approximation for ``optimism'' by treating arms which have been played less as more uncertain (and thus plausibly better) than arms that have been played frequently. In UCB1's strict formulation, the bound is derived from the Chernoff-Hoeffding bound \cite{Chernoff:1952,Hoeffding:1963,Bentkus:2004} on the right tail distributions for the estimation of Bernoulli random variables, but the confidence bound model applies equally well to any distribution where an appropriate bound can be defined.

The second term in the maximization criterion has been extended, as in the MOSS algorithm \cite{Audibert/Bubeck:2009}  (discussed in an upcoming section) to consider the remaining horizon to create an ``exploratory value'' that is declining in finite time or to improve the tightness of the bound on variance.

UCB1 as specified has a strict bound on regret at time $t$, for Bernoulli arms, given by the following formula, shown in the original paper,

\begin{equation}
8 \cdot \left[{\sum_{i:\mu_i < \mu^*} \left(\frac{\ln t}{\Delta_i}\right)}\right] + \left(1 + \frac{\pi^2}{3}\right)\left(\sum_{i=1}^K \Delta_i\right).
\end{equation}

\subsection{UCB2}
UCB2, an iterative improvement over UCB1, reduces the constant term in the fraction of time a suboptimal arm will be selected, reducing the overall regret, at the cost of only a slightly more complicated algorithm.

In UCB2, iterates are broken into epochs of a varying size. In each epoch, arm $i$ is selected to maximize $\bar{x}_i + \sqrt{\frac{(1 + \alpha)(\ln(en / (1+\alpha)^{r_i}))}{2(1+\alpha)^{r_i}}}$ and then played exactly $\ceil{(1 + \alpha)^{r_i + 1} - (1+\alpha)^{r_i}}$ times before ending the epoch and selecting a new arm. $r_i$ is a counter indicating how many epochs arm $i$ has been selected in and $0 < \alpha < 1$ is a parameter that influences learning rate discussed below.

The finite-time bound on regret for UCB2 is known for times $t \geq \max_{i:\mu_i < \mu^*} \frac{1}{2\Delta_i^2}$ and is given by,

\begin{equation}
\sum_{i: \mu_i < \mu^*} \left(\frac{(1+\alpha)(1+4\alpha)\ln(2e\Delta_i^2t)}{2\Delta_i} + \frac{c_\alpha}{\Delta_i}\right)
\end{equation}

where $e$ is Euler's constant and $c_\alpha = 1+\frac{(1+\alpha)e}{\alpha^2} + (\frac{1+\alpha}{\alpha})^{(1+\alpha)} (1 + \frac{11(1+\alpha)}{5\alpha^2 \ln(1+\alpha)})$ as proven in \cite{Auer/Cesa-Bianchi/Fischer:2002}. The important property of $c_\alpha$ to notice is that $c_\alpha \rightarrow \infty$ as $\alpha \rightarrow 0$, forcing a trade-off between the selection of $\alpha$ to minimize the first term towards $1/(2\Delta_i^2)$ and the second term. The original paper suggests optimal results from setting $\alpha$ such that it is decreasing slowly in $t$ but is not specific to the form of decrease; in practice, they also demonstrate, the choice of $\alpha$ does not seem to matter much as long as it is kept relatively small.

\subsection{UCB-Tuned}
A strict improvement over both UCB solutions can be made by tuning the upper-bound parameter in UCB1's decision rule. Specifically, \cite{Auer/Cesa-Bianchi/Fischer:2002} further expands these solutions by replacing the second term $\sqrt{\frac{2 \ln t}{n_i}}$ with the tuned term $\sqrt{\frac{\ln t}{n_i} \min({\frac{1}{4}, V_i(n_i))}}$ where $V_i$ is an estimate of the upper-bound of the variance of arm $i$ given by, for example,

\begin{equation}
V_i(n_i) \equiv \left(\frac{1}{n_i}\sum_{\tau=1}^{n_i} X^2_{i,\tau}\right) - \bar{X}^2_{i,n_i} + \sqrt{\frac{2\ln t}{n_i}}
\end{equation}

where $n_i$ is the number of times arm $i$ has been played out of $t$ total plays. UCB-Tuned empirically outperforms UCB1 and UCB2 in terms of frequency of picking the best arm. Further, \cite{Auer/Cesa-Bianchi/Fischer:2002} indicate that UCB-Tuned is ``not very'' sensitive to the variance of the arms. Simple experimentation shows that UCB-Tuned as defined above outperforms the earlier UCBs significantly in our results and in the original paper's exposition.

\subsection{MOSS}
MOSS \cite{Audibert/Bubeck:2009}, or the Minimax Optimal Strategy in the Stochastic case, produces a variant of UCB1 that is presented in a generalized context, such that it can apply to all known bandit variants or subproblems. In MOSS, the $\ln{t}$ component of the padding function in UCB1 for arm $i$ is replaced with $\ln{\frac{H}{Kn_i}}$ where $n_i$ is the number of times arm $i$ has been played, $H$ is the total number of iterates to be played (the horizon, at the beginning) and $K$ is the number of arms available in a stochastic (non-adversarial) bandit problem. The work of \cite{Audibert/Bubeck:2009} shows that expected regret for MOSS is bounded from above, by,

\begin{equation}
	\mathbb{E}R \leq 25 \sqrt{HK} \leq \frac{23K}{\Delta} \log\left(\max\left(\frac{140H\Delta^2}{K}, 10^4\right)\right)
\end{equation}

where $\Delta = \min_{i : \Delta_i > 0} \Delta_i$, the smallest gap between the optimal arm and the second best arm. Note that this calculation of regret applies continuously in the stochastic case, but we will see later in the adversarial discussion that it is complicated in that environment due to non-unicity of the optimal arm.

\subsection{KL-UCB}
KL-UCB \cite{Maillard/Munos/Stoltz:2011} presents a modern approach to UCB for the standard stochastic bandits problem where the padding function is derived from the Kullback-Leibler (K-L) divergence. KL-UCB demonstrates regret that improves the regret bounds from earlier UCB algorithms by considering the distance between the estimated distributions of each arm as a factor in the padding function. Specifically, define the Kullback-Leibler divergence \cite{Kullback/Leibler:1951, Garivier/Cappe:2011} (for Bernoulli distribution arms\footnote{For non-Bernoulli arms, the more generic formula $\int_{-\infty}^{\infty} p(x) \log \frac{p(x)}{q(x)} dx$ should be used, where p and q are distribution densities.}) as,

\begin{equation}
\label{KLdiv}
d(p,q) = p \log \frac{p}{q} + (1-p)\log\frac{1-p}{1-q}
\end{equation}

with convention of $0\log 0 = 0$, $0\log\frac{0}{0} = 0$, and $x \log \frac{x}{0} = +\infty$ for $x>0$. The Kullback-Leibler divergence $d(p, q)$ provides a probability-weighted measure of the difference between two distributions which does not rely on collapsing the distribution to a midpoint (e.g., expectation). 

To pick an arm in each iteration of KL-UCB, we maximize

\begin{equation}
i = \argmax_i ~ (n_i \cdot d(\mu_i, M)) \leq \log t + c \log\log t
\end{equation}
 
where $M$ is picked from the set of all possible reward distributions. The K-L divergence of $d(x, M)$ is strictly convex and increasing in $[x,1)$ \cite{Garivier/Cappe:2011} making this equation efficiently computable by solving the convex optimization problem.

\subsection{Bayes-UCB}

Alongside KL-UCB, Bayes-UCB \cite{Kaufmann/Cappe/Garivier:2012} --- an explicitly Bayesian variant of UCB --- represents the current state of the art of UCB algorithms. It is an asymptotically efficient advanced algorithm with empirical results which outperforms KL-UCB. In the Bayesian approach to the multi-armed bandit problem, each arm is represented as an estimate of a distribution that is updated in the traditional Bayesian fashion and the decision is selected from the bandit which has the highest \emph{score}. The score is captured as a dynamic in time quantile of the posterior estimate. \cite{Kaufmann/Cappe/Garivier:2012} show that this Bayesian-derived UCB has a cumulative regret that empirically outperforms the strongest of the original UCB algorithms by a substantial margin in a handful of selected problems while having the advantage of being distribution agnostic and showing the early-iterate flexibility of a Bayesian approach to knowledge acquisition. A computational complexity challenge is acknowledged but not explored in depth. 

Bayes-UCB is similar to the \emph{probability matching} strategies to be discussed later: quantiles of a distribution are estimated to increasingly tight bounds and the probability of a given arm ``being the best'' is used to determine the next step. To perform Bayes-UCB, the algorithm requires a prior on the arms, $\Pi^0$ and a function to compute the quantiles of the expected distributions, $Q(\alpha, \rho)$ such that $P_\rho(X \leq Q(\alpha, \rho)) = \alpha$. At each time step $t$, Bayes-UCB draws the arm $i$ to maximize the quantile selected as follows. It picks

\begin{equation}
i = \argmax_i q_i(t) = Q(1 - \frac{1}{t}, \lambda_i^{t-1})
\end{equation}

where $Q$ meets the property described above and $\lambda_i^{(t-1)}$ is the estimated posterior distribution of arm $i$ at the previous time step. This is then updated according to the Bayesian updating rule and used as the prior for the next iteration. 

In a theoretical analysis, \cite{Kaufmann/Cappe/Garivier:2012} show that Bayes-UCB achieves asymptotic optimality and a non-asymptotic finite-time regret\footnote{Specifically, they show this for the variant where Q is computed with a horizon dependence $Q(1 - \frac{1}{t(\log H^c)}, \lambda_j^{t-1})$. We present the simpler variant $Q(1 - \frac{1}{t}, \cdot)$ because in simulations $c=0$ demonstrated the best results. Derivation of asymptotic optimality is dependent on the horizon term, though this appears to largely be a mathematic artifact.} in $O(H)$.

It is interesting to note that by treating the quantile function and underlying model appropriately, Bayes-UCB can, in theory, represent any distribution and most subproblems of the multi-armed bandit. As a simple but valuable example, by representing the underlying model as a Bayesian regression, one can include contextual information in the bandit process or even the weighted least squares decay process discussed for nonstationary bandits in a later section.

\subsection{POKER and price of knowledge}	

A non-UCB algorithm, POKER \cite{Vermorel/Mohri:2005} or Price of Knowledge and Estimated Reward is a generalizable economic analysis inspired approach to the problem. 

The intuition behind POKER is to assign a ``value'' of the information (the ``exploration bonus'') gained while pulling a given arm. This value is estimated for each arm, and then the arm with the highest expected payoff \emph{plus} expected value of information is played. Value of information is defined to maximize expected outcome over the horizon. To gain an intuition, first assume an oracle provides the best arm as arm $i^*$ with payoff $\mu^*$, that we have an estimate for each arm's payoff $\hat{\mu}_i$ and that we have an estimated best arm $\hat{i}$ with estimated payoff $\hat{\mu}^*$. Define the magnitude of the expected improvement as $\delta = E[\mu^* - \hat{\mu}^*]$, then the probability of an improvement for a given arm is $P[\mu_i - \hat{\mu}^* \geq \delta]$.

When there are $(H-t)$ plays left, any new knowledge found in this iterate can be exploited $(H-t)$ times. This means the expected improvement has a (non-discounted) value of $\delta \cdot (H-t)$. 

A problem arises in computing $\delta$, as if $i^*$ and $\mu^*$ were known, there would be no need to explore. Instead, the ordered estimate of the means are used. Imagine, an ordered list of the empirical mean rewards as $\hat{\mu}_{i_1} \geq \cdots \geq \hat{\mu}_{i_K}$. \cite{Vermorel/Mohri:2005} choose, based on empirical results, to approximate $\delta$ proportional to the gap between $\hat{\mu}_{i_1}$ and the $\hat{\mu}_{i_{\sqrt{K}}}$ arm. Specifically, they set $\delta = \frac{\hat{\mu}_{i_1} - \hat{\mu}_{i_{\sqrt{K}}}}{\sqrt{K}}$. That is, if there are $K$ arms, the difference between the best and the $\sqrt{K}$th best current estimate is proportional to the plausible gain.

In the limit (as the number of arms approaches infinity), this approximation strategy ensures bias and variance minimization.

Additionally, one can observe that the whole probability $P[\mu_i - \hat{\mu}^* \geq \delta] = P[\mu_i \geq \hat{\mu}^*  + \delta]$ is approximated (or identical, in the event of normally distributed means\footnote{This is true in the limit by the central limit theorem, but as there may be a small number of arms and trials, it may be a poor approximation in some environments.}) by the cumulative probability of the reward \emph{higher} than the best empirically expected reward plus expected improvement $\hat{\mu}^* + \delta$,

\begin{equation}
P[\mu_i \geq \hat{\mu}^* + \delta] = \int_{\hat{\mu}^*+\delta}^{\infty} \mbox{N}\left(x, \hat{\mu}_i, \frac{\hat{\sigma}_i}{\sqrt{n_i}}\right) dx
\end{equation}

where $N(x, \mu, \sigma)$ represents the normal distribution and $n_i$ means the number of times arm $i$ has been played and mean $\mu_i$ and variance $\sigma_i$ take on their usual meaning.

This gives us sufficient information to define a decision criterion. Select the arm which maximizes the expected sum of total rewards over the horizon $H$. Formally, at each time step $t$, select arm $i$ to play:

\begin{equation}
\argmax_i \mu_i  + \delta (H-t) P[\mu_i \geq \hat{\mu}^* + \delta].
\end{equation}

Note that $\delta(H-t)$ is the total expected gain over the remaining horizon. By multiplying by the probability this arm will actually exceed the best known arm, we achieve a sensible expectation to maximize. This value could be easily time-discounted by introducing a sum of discounted payoffs if the time horizon was measured at a scale where time-discounting were of value.

POKER uses knowledge of the length of the horizon or number of plays that remain, $(H-t)$, as a parameter that effectively determines how to weight exploration and exploitation. The authors make the claim that requiring the horizon explicitly is a more intuitive parameter than the parameters associated with many other algorithms. Additionally, the parameter can be set to a fixed value to simply use it to balance exploration and exploitation in the case horizon is unknown or infinite. 

\cite{Vermorel/Mohri:2005} introduce the POKER policy and use the term \emph{zero-regret strategy} to describe it. In their context, \emph{zero-regret} means guaranteed to converge on the optimal strategy, eventually: that is, a strategy which has average per-play regret tending to zero for any problem which has a horizon tending to infinity. The term \emph{zero-regret} will not be used in the rest of our discussion, preferring instead ``guaranteed to converge to zero.''

The authors compare POKER to $\epsilon$ strategies, Exp3 (discussed in a future section) and others on a real world redundant retrieval\footnote{The problem is to identify the fastest source for a content delivery network with numerous redundant sources of requested data.}  routing problem and find that POKER outperforms $\epsilon$ strategies by a factor of approximately 3. As of this writing, there has been no known finite time analysis of regret for POKER. 

\subsection{Thompson Sampling and Optimism}
While POKER provides a strong generalization of many of the problems faced in adaptive experiment design with respect to value of information, recently, a simpler technique -- Thompson sampling -- has been shown to perform competitively to the state-of-the-art in a variety of bandit and other learning contexts. Thompson \cite{Thompson:1933}, ``\emph{On the likelihood that one unknown probability exceeds another in view of the evidence of two samples}'' produced the first paper on an equivalent problem to the multi-armed bandit in which a solution to the Bernoulli distribution bandit problem now referred to as \emph{Thompson sampling} is presented. 

The stochastic solution presented by \cite{Thompson:1933} involves \emph{matching} the probability of playing a particular arm with the arm's inherent ``probability of being the best'' given the data observed by sampling from each distribution precisely once and selecting the maximum sample. The language \emph{probability matching} arises from this intuition and seems to originate from \cite{Morin:1955}. \emph{Probability matching} is extensively used in the experimental psychology literature to describe the behavior matching action probabilities to the probability of an outcome. This concept is distinct from the actual implementation of sampling precisely once from the posterior estimate to simulate the optimality pseudo-distribution, which we refer to as Thompson sampling. A factor motivating this interplay of nomenclature is the increasingly common use of multi-armed bandit processes in the modelling of animal and human psychology and behavior \citep[e.g.,][]{Srivastava:2013,Reverdy:2014}.

\cite*{Scott:2010} applies a strictly Bayesian framework to presenting Thompson sampling and specifically calls it \emph{randomized probability matching}. We will simply use the language of \emph{Thompson sampling} through the rest of this discussion.

Recent research in Thompson sampling has provided an information-theoretic analysis drawing ideas similar to the ``price of knowledge'' \cite{Russo/VanRoy:2014}, various proofs (including and demonstrations of regret minimization \cite{Guha/Munagala:2014,Gopalan/Mannor/Mansour:2014}, a technique to apply Thompson sampling via the online bootstrap \cite{Eckles/Kaptein:2014}, exploration of the cold-start problem\footnote{The cold-start problem is a particularly apt use of bandit modelling. Specifically, the problem models the issue of being unable to draw any recommendations for new users until sufficient data has been collected for said user to fit an appropriate model or prediction to his or her preferences. The multi-armed bandit model provides exploratory guidance in some contexts to help address this problem.}  in recommender systems \cite{Nguyen/Mary/Preux:2014} and numerous applications of the technique \cite{Bouneffouf/Laroche/Urvoy/Feraud/Allesiardo:2014,Gopalan/Mannor/Mansour:2014,Neufeld/Gyorgy/Schuurmans/Szepesvari:2014,Shahriari/Wang/Hoffman/Boucard/deFreitas:2014}.

Strict bounds on regret were a hindrance to theoretical adaption of generalized Thompson sampling, however, recently, bounds for a specific models (e.g., traditional K-armed bandits with beta prior distributions) have been discovered by \cite*{Agrawal/Goyal:2012,Kaufmann/Korda/Munos:2012,Russo/VanRoy:2014}. For $K=2$, their bound on regret is given as $O(\frac{\ln T}{\Delta})$; for $K>2$ the bound is significantly less tight, as $O\left(\frac{\ln T}{{\sum_{i=2}^K {({\Delta_i}^2})}^2}\right)$. Significantly, the information-theoretic work of \cite{Kaufmann/Korda/Munos:2012,Russo/VanRoy:2014} proves efficient ($O(\log H)$) regret bounds for Thompson sampling and show convincingly that Thompson sampling performs comparably to a correctly-tuned UCB-type algorithm in general. This is a result which had been expected, however is significant as Thompson sampling is a more general solution than any particular implementation of a UCB-type algorithm.

In order to formalize the matching of our play probabilities with the probability of a given play being the best play, we adopt a Bayesian framework\footnote{The interplay between frequentist methods, Bayesian methods and Thompson sampling is discussed in depth in \cite{Kaufmann/Cappe/Garivier:2012}.} and, in general, a parametric distribution over a parameter set $\theta$.  We can compute the probability at time $t$ of a given arm providing optimal reward as,

\begin{equation}
\int \mathbbm{1}  \left[S_t = \argmax_{i=1,...,K} E_\theta[x_{i,t}]\right] P(\theta | x) d\theta.
\end{equation}

Rather than computing the integral, \cite{Thompson:1933} and others show that it suffices to simply sample from the estimated payoff distribution at each round and select the highest sampled estimated reward. That is, the repeated selection of the maximum of a single draw from each distribution, produces an estimate (and thus selection behavior) of the \emph{optimality distribution}. This result, while long known, is surprising and valuable, turning an intractable problem in to a computationally simple one.

\subsubsection{Optimism and Confidence Bounds in Probability Matching}

A recurring perspective on the efficient use of uncertainty within such a multi-armed bandit (and exploratory learning in general) has been that of ``optimism in the face of uncertainty'' \cite{Lai/Robbins:1985,Auer/Cesa-Bianchi/Fischer:2002,Szita/Lorincz:2008,Munos:2014}. The idea is presented as a method for treating uncertainties and balancing exploration: when a statistical uncertainty is present, a small but consistent gain in outcome \cite{Chapelle/Li:2011} can be achieved by simply remaining optimistic and assuming the value is in the ``more desirable'' portion of the distribution under uncertainty. 

This idea has been seen already in many of the static (non-probability matching) algorithms presented prior. For example, any UCB-type algorithm derives its action estimates from an ``optimistic'' surrogate about the state of the empirical estimate. This form of static optimism is the basis of most algorithms for multi-armed bandits, though the mechanism for defining optimism is variable.

Optimism improves the result of the Thompson sampler \emph{in terms of regret}. In an efficient data-collection or experiment design paradigm, one may wish to improve the quality or confidence of their estimation, in this environment an optimism benefit has not been explored. Kaptein \cite{Kaptein:2015} presents an argument in favor of using pure Thompson sampling for optimal experiment design showing that the same variance-weighting property which makes non-optimistic Thompson sampling favorable for regret minimization provides efficient exploration for capturing effect sizes in terms of data collection. 

\subsection{Best Empirical Sampled Average (BESA)}
Baransi et al. \cite{Baransi/Maillard/Mannor:2014} present an algorithm called BESA which provides a fully nonparametric sampling strategy built upon the intuitions of Thompson sampling and the sub-sampling literature \cite{Romano/Shaikh:2012}. BESA shows empirical performance comparable to properly tuned Thompson samplers and other state-of-the-art algorithms. Being a fully nonparametric strategy, BESA provides a significant advantage in simplicity and robustness over traditional Thompson sampling: the same implementation can be used for any implementation \emph{and} knowledge of the underlying reward distribution for any given implementation is entirely unnecessary. This strategy comes at a cost, with their analysis showing 2-20x more computation time for a naive implementation of BESA than the very computationally-efficient Thompson sampler.

In the two-arm case, the algorithm proceeds simply by acknowledging that it isn't ``fair'' to directly compare the empirical means between two arms with different numbers of observations. For example, if arm $1$ (let $n_1$ represent the number of times $n_1$ was played) had been observed substantially more frequently than arm $2$ ($n_2$, respectively), the estimated means do not adequately account for the greater uncertainty in the smaller sample set. To compensate for this, BESA uses a uniform sampling strategy, sampling $n_2$ observations from the observed rewards of arm $1$, then comparing the empirical average of the rewards from arm $2$ to the empirical average of the subsampled rewards from arm $1$, with the sizes of the two sets now being equal. Ties in the averages are to be broken in favor of the less played arm.

To extend this beyond the two-arm case, the authors suggest a divide-and-conquer tournament strategy which randomly reorders the arms and then performs a recursive halving of the set of arms until only two arms to be compared are left in each node of the tree. The two arms in each node are run through the two-arm case and the winner is returned to the prior level. Formally, this corresponds to a binary tree of height $\lg K$ with each level representing a round of the tournament and the root node containing the singular winner.

The authors present an analysis showing an $O(\log H)$ finite-time regret for BESA. 

\section{Complications of the Simple Stochastic Model}
Thus far we have presented a discussion of algorithms for the simplest variant of the problem. In many contexts, algorithms which are efficient for the simple stochastic model are not efficient when a complication has been introduced. The most general of those is the adversarial bandit which \cite{Audibert/Bubeck:2010} provide a framework from which many of the other variants can be derived. The contextual bandit is perhaps the most interesting from an experiment design perspective, allowing the introduction of other dimensions which plausibly covary with the selected treatment to produce a given effect. The ability to measure these covarying parameters in an online fashion is extremely valuable from an efficient design perspective. Other variants we explore include the nonstationary environment, appropriate for tracking many forms of contextual misspecification and long-running or infinite-horizon experiments, and multiple simultaneous experimentation with bandits with multiple plays.
\subsection{Adversarial Bandits}
One of the strongest generalizations of the k-armed bandit problem is the adversarial bandits problem. In this problem, rather than rewards being picked from an \emph{a priori} fixed distribution, rewards are selected, in the worst case per-play, by an adversary. The problem is transformed into an iterated three step process; in step 1, the adversary picks the reward distributions (generally with full availability of the list of prior choices, though the constraints on the distribution are discussed); in step 2, the agent picks an arm without awareness of the adversary's selections; in step 3, the rewards are assigned. This is a strong generalization because it removes the distribution dependence on the arms (and as such, stationarity and other distribution-dependent assumptions); an algorithm that satisfies the adversarial bandits problem will satisfy more specific\footnote{Especially, contextual bandits.} bandits problems, albeit, often sub-optimally. To state the model succinctly, adversarial bandits are multi-armed bandit problems in which there is no assumption of a statistical reward generating process. Any strategy for the adversarial bandit problem must acknowledge the potential \emph{information asymmetry} between the player and the adversary. 

Without constraints, the adversarial bandit can be seen as a competition between the algorithm and a omniscient adversary with unlimited computational power and memory capable of always staying ahead of any strategy the agent selects. As adversarial bandits are such a strong generalization, \cite*{Audibert/Bubeck:2010} provide a taxonomy of bandit problems that builds from the constraints on the adversary's selection process. Fundamentally, allowing the distribution to vary in each time, they let $n$ represent the number of possible distributions available. They then provide five distinctions. (1) The purely \emph{deterministic bandit problem}, where rewards are characterized as a matrix of $nK$ rewards, where $K$ represents the number of arms and $n$ the number of time steps. In each time step, a single deterministic reward is set (fixed \emph{a priori}) for each arm. (2) The \emph{stochastic bandit problem} -- the variant discussed in the majority of this work -- in this taxonomy is characterized by a single distribution for each arm, stationary in time, independent and bounded on some range, say, $x_i \in [0,1]$. (3) The \emph{fully oblivious adversarial bandit problem}, in which there are $n$ distributions for each arm, independent of each other (both through time and across arms) and independent of the actor's decisions, corresponding to changes selected by the adversary across time. (4) The \emph{oblivious adversarial bandit problem}, in which the only constraint is that the distributions are selected independent of the actor's decisions. Finally, (5) the adversarial bandit, in their work referred to as the \emph{non-oblivious bandit problem}, where the reward distributions can be chosen as a function of the actor's past decisions.

In the majority of this work, we focus explicitly on the stochastic variants of the multi-armed bandit problem, choosing a lens by which deterministic or even simply non-oblivious bandits are not known to be deterministic or non-oblivious by the agent ahead of time. Our lens models the various forms of oblivious bandits as considerations to the stochastic nature of the problem, for example, treating contextual covariates and non-stationarity as a form of statistical misspecification, even when sometimes that misspecification will be impossible to resolve (as in the case of \emph{Knightian uncertainty} \cite{Knight:1921}, where the correct reward model has some immeasurable and incalculable component). This differs from the lens in \cite*{Audibert/Bubeck:2010}, providing a prospective which applies more closely to the application area of interest in this work (one in which the true underlying model almost certainly consists of unknown or unknowable covariates, but is also partially approximated by variables we can observe), but comes at the cost of not generalizing to the pure non-oblivious adversarial problems.

The major caveat of adversarial bandits, is that our definition of ``performance'' needs to be relaxed for any measures to be meaningful. Specifically, a strong performing algorithm must be defined using a measure of regret that compares our decisions solely to a fixed machine over time, that is, a strong adversarial bandit can still achieve logarithmic regret, but only if the ``best arm'' is defined at time $t=0$ and does not vary across trials. To rephrase, that means that of our definitions of regret given earlier in this chapter, only the ``weak-regret'' notions can be meaningful within an adversarial context.


The majority of the efficient solutions to adversarial problems are variants of the Exp3 algorithm presented in \cite*{Auer/Cesa-Bianchi/Freund/Schapire:2002} for the general, no statistical assumptions adversarial bandits case. \cite*{Beygelzimer/Langford/Li/Reyzin/Schapire:2010} extend the work of \cite{Auer/Cesa-Bianchi/Freund/Schapire:2002} and \cite{McMahon/Streeter:2009} to transform Exp4 to produce a high-probability bounded version called Exp4.P.
\subsubsection{Hedge and Exp3}

\cite{Auer/Cesa-Bianchi:1998} present the first look at the adversarial bandit problem and include an algorithm with high-probability bounded regret called Exp3: the \textbf{exp}onential-weight algorithm for \textbf{exp}loration and \textbf{exp}loitation based on an algorithm called Hedge for the full information problem. Exp3 \cite{Auer/Cesa-Bianchi/Freund/Schapire:2002} presents a readily understandable, simple algorithm for adversarial bandits. Given a pure exploration parameter, often called \emph{egalitarianism}, $\gamma \in [0, 1]$, which measures the fraction of time the algorithm selects a purely random decision, the algorithm then spends $(1-\gamma)$ of the time doing a weighted exploration/exploitation based on the estimated actual reward.

The estimation process for Exp3 is an exponentially updating probability weighted sample. The arm weight is updated immediately after pulling a given arm and being delivered the reward $\rho_i$ with the formula

\begin{equation}
w_{i,t} = w_{i,t-1} \cdot e^{\gamma \cdot \frac{\rho_i}{p_{i,t}\cdot K}}
\end{equation}

where $w_{i,t}$ is the arm $i$ specific weight at time $t$ and $p$ is our selection criteria. The probability of each specific arm to play in each iteration is selected according to $p$, which considers the arm weighting and $\gamma$ semi-uniformity, namely,

\begin{equation}
p_{i,t} = (1-\gamma) \frac{w_{i,t}}{\sum_{j=1}^K w_{j,t}} + \gamma \cdot \frac{1}{K}.
\end{equation}

In some sense, Exp3 combines the semi-uniformity in the  $\epsilon$ strategies with an exponentially decayed  ``probability of best'' weighted exploration/exploitation similar to probability matching methods. 

A computationally efficient version of Exp3 called Exp3.S is presented in \cite*{Cesa-Bianchi/Lugosi:2006}. 

\subsubsection{Exp4}

Exp3 does not include any concept of contextual variables or ``expert advice''. \cite{Auer/Cesa-Bianchi/Freund/Schapire:2002} develop an extension of Exp3, called Exp4 (Exp3 with \textbf{exp}ert advice). Exp4 is identical to Exp3, except the probability of play is selected with the addition of a set of $N$ context vectors $\xi$ per time and the weight function is similarly replaced to consider the context vectors. One should note that the weights $w$ are now computed per context vector, where a context vector can be viewed as an ``expert'' advising of a selection coefficient for each arm; we now use $j$ to indicate the index of the expert and continue to use $i$ to indicate the index of the arm, for clarity,

\begin{equation}
\label{policy:exp4:weight}
w_{j,t} = w_{j,t-1} \cdot e^{\gamma \cdot \frac{\rho_j \cdot \xi_{j, t}}{p_{j,t} \cdot K}}.
\end{equation}

For the selection probability $p$, interpret $\xi_{j,t}(i)$ as the advice coefficient expert $j$ gives at time $t$ about arm $i$,

\begin{equation}
p_{i,t} = (1-\gamma) \sum_{i=1}^N \frac{w_{j,t} \xi_{j,t}(j)}{\sum_{k=1}^K w_{k,t}} + \gamma \cdot \frac{1}{K}.
\end{equation}

Note that $k$ represents an iterator over all arms in the second term. With a minor abuse of notation, this is equivalent to Exp3 where we update our weight vector with the context $\xi$, reward $\rho$, and selection probability $p$ according to $\xi \cdot \rho/p$ for the arm played at each time step except that the weight vector is now the summed contextual weight vector.

\paragraph{Exp4.P}

Exp4.P is a variant of the Exp4 algorithm presented in \cite*{Beygelzimer/Langford/Li/Reyzin/Schapire:2010} with asymptotically bounded\footnote{The notation $\tilde{O}(n)$ is read ``soft-O of n'' and is equivalent to $O(n \log^k n)$, i.e., the big-O notation where logarithmic factors are ignored.} regret in the high-probability case of $\tilde{O}(\sqrt{KH \log N})$ where $K$ and $H$ take their usual meaning and $N$ is the number of context vectors as indicated in the previous section. The bound does not hold in the original Exp4 presentation \cite{Lazaric/Munos:2009}, as the variance of importance-weighted numerator term is too high \cite{Beygelzimer/Langford/Li/Reyzin/Schapire:2010}. Exp4.P modifies Exp4 such that the bound holds with high-probability. The change in Exp4.P is only in how the weight vector is updated at time $t$. Rather than using Equation (\ref{policy:exp4:weight}), Exp4.P uses an updating function,

\begin{equation}
w_{j,t} = w_{j,t-1} \cdot e^{\frac{\gamma}{2} \cdot \left(\rho_j \cdot \xi_{j,t} + \hat{v}_{j,t} \sqrt{\ln{(N/\delta)}/KH}\right)}
\end{equation}

where $\delta>0$ is a parameter that defines the desired probability bound of the regret ($1-\delta$) and $v_{j,t}$ is defined as

\begin{equation}
v_{j,t} = \sum_{1, ..., K} \frac{\xi_{i,t}(j)}{p_{i,t}}.
\end{equation}

This modification allows \cite*{Beygelzimer/Langford/Li/Reyzin/Schapire:2010} to bound regret of the new algorithm, Exp4.P, with probability of at least $1-\delta$ to $-6\sqrt{KH \log (N/\delta)}$.

\subsubsection{Stochastic and Adversarial Optimal (SAO)}
\cite*{Bubeck/Slivkins:2012} introduce a testing technique that is capable of handling both the stochastic (non-adversarial) problem and the adversarial problem with near-optimal regret results. Stochastic problems generally use a different definition of regret than adversarial problems, so the analysis provided in this work takes place in two parts assuming the model is \emph{either} stochastic \emph{or} adversarial showing asymptotically regret of $O(\mbox{polylog}(n))$ in the stochastic case\footnote{The notation $O(\mbox{polylog}(n))$ means $O((\log n)^k)$ for some $k$. This is similar to the use of $\tilde{O}$ to indicate the insignificance logarithmic terms often bring to the analysis of algorithms.} and the $O(\sqrt{n})$ pseudo-regret from Exp3 in the adversarial case.

SAO proceeds in three phases, making it a semi-uniform strategy: exploration, exploitation and the adversarial phase. The exploration and exploitation phases are largely as expected, interleaved to operate pairwise (arm 1 vs. arm 2) and rule out ``suboptimal'' arms as it progresses. For the remainder of this exposition, assume there are only two arms and arm 1 is strictly superior to arm 2. Further, let $C \in \Omega({\log n})$ be an arbitrary parameter which enforces consistency, selected specifically for the application area, for example $C=12 \log(n)$, let $\tilde{H}_{i,t}$ be the average observed reward for arm $i$, $t$ represent time (number of iterates so far) and $\tau_*$ represent the point we switch from exploration to exploitation.

We start in a state of exploration, where we pick an arm with equal probability for a minimum of $C^2$ rounds and until we find a ``sufficiently superior'' arm according to the following condition:

\begin{equation}
|\tilde{H}_{1, t} - \tilde{H}_{2,t}| < \frac{24 C}{\sqrt{t}}
\end{equation}

During the exploitation phase, the arms are drawn according to the probabilities $p_t(2) = \frac{\tau_*}{2t}$ and $p_t(1) = 1-p_t(2)$, that is, the probability of drawing the suboptimal arm is decreasing asymptotically in time. A set of conditions is checked to see if the observed rewards still fit within the expected stochastic model. The conditions checked are referred to as \emph{consistency conditions} and are as follows.

The first consistency condition, which checks if the observed rewards in exploitation are congruent with the findings of the exploration phase, that is, whether the rewards are bounded in a range consistent with our observation that arm 1 is better than arm 2 by approximately the observed amount. Concretely, the first consistency condition is 
\begin{equation}
\frac{8 C}{\sqrt{\tau_*}} \leq \tilde{H}_{1,t} - \tilde{H}_{2,t} \leq \frac{40 C}{\sqrt{\tau_*}}.
\end{equation}

The second consistency condition, which checks that arm $i$'s estimate is still within bounds of the expected estimate, consistent with the fact that during exploitation the suboptimal arm is drawn with low probability. Consider $\hat{H}_{i,t}$ to be the expected reward from arm $i$ at time $t$ given that the world is stochastic and the arm can be appropriately modelled, so that $\tilde{H}_{i,t} - \hat{H}_{i,t}$ represents the difference in the expected reward and the observed reward. Concretely, the second consistency conditions are,
\begin{equation}
|\tilde{H}_{1,t} - \hat{H}_{1,t}| \leq \frac{6 C }{\sqrt{t}},
\end{equation}
\begin{equation}
|\tilde{H}_{2,t} - \hat{H}_{2,t}| \leq \frac{6 C }{\sqrt{\tau_*}}.
\end{equation}

All the \emph{magic numbers} in these conditions are derived from the high-probability Chernoff bounds for the stochastic case. The different denominators on the right hand side of the equation account for the low probability of drawing the inferior arm (arm $2$) during exploitation.

In the event any of the consistency conditions fail, we assume the model is non-stochastic and switch from the explore-exploit algorithm to that of Exp3. The work explores and proves properties of the conditions. Selection of the consistency parameters is important, as they would allow a carefully crafted adversary to maintain the conditions. Such conditions cannot allow the adversary to create a high level of regret for the application yet must hold in high probability in the non-adversarial case.

This algorithm as described combines the asymptotic regret bounds of both UCB1 and Exp3 in a near-optimal (asymptotic) fashion for both stochastic and the most general form of adversarial bandits. There is no analysis of the finite time regret.

\subsection{Contextual Bandits for  Experiments with Covariates}

The simple $k$-armed bandit problem performs sub-optimally by its design in the advertising context. In general, the contextual bandits framework is more applicable than the non-contextual variants of the problem, as it is rare that no context is available \cite{Langford/Zhang:2007}\footnote{While it is rare that no context is available, it is \textbf{not} rare that the value of the context is entirely unknown -- in the stylized example of a slot machine, the arms may be different colors, whether that is a determining factor in the payoff probabilities or not may be \emph{a priori} completely unknowns.}. Specifically, the simplest form of the model selects from $k$ advertisements then discovers the payout associated with that particular play.  A similar consideration may improve results in the clinical environment when considering factors of the patient or treatment as contextual variables that are expected to interact with the treatment itself in determining its reward.

The contextual bandit setting has taken many names including bandits with context, bandit problems with covariates \cite{Woodroofe:1979,Sarkar:1991,Perchet/Rigollet:2013}, generalized linear bandits, associative bandits \cite{Strehl/Mesterharm/Littman/Hirsh:2006} and bandit problems with expert advice \cite{Auer/Cesa-Bianchi/Fischer:2002}. The contextual bandit problem is closely related to work in machine learning on supervised learning and reinforcement learning; indeed, some authors \cite{Dudik/Hsu/Kale/Karampatziakis/Langford/Reyzin/Zhang:2011} have referred to it as ``the half-way point'' between those fields because of the ability to construct algorithms of a reinforcing nature with convergence guarantees while considering relatively general models.

We further divide the context into both the arm-context (such as properties of the selection variable, for instance, dosage) and world-context being selected for (such as properties of the patient or testing environment). Arm-context can be used to learn shared properties across arms, while world-context interacts with arm context and is declared on a per step basis. A more complicated hierarchical model could be envisioned where data is available at many levels of granularity, but little work exists to support this form of analysis in multi-armed bandits. Contextual variables allow a much more rich learning process where the vector of contextual variables can be used to guide learning, even if they are incomplete, and in general do not significantly harm the learning process if they are not strongly covariant to the variable of interest.

Broadly, the expected reward can be approximated by a model of the form

\begin{equation}
Y_i = \alpha + \beta A_i + \gamma W_t + \xi A_i W_t + \epsilon
\end{equation}

where $Y_i$ indicates the expected payoff of a given arm conditional on the context, $\beta$ indicates the coefficient vector as a function of the arm context, and $\gamma$ a coefficient vector of the world context. In the web search context, the world context vector might be the words included in the search query, in which case we would expect our agent, in the limit, to learn a model that suggests ideal advertisements related to the query for any given search.

A slightly more general form of contextual or side-information bandits is referred to as associative reinforcement learning \cite{Strehl/Mesterharm/Littman/Hirsh:2006} in some statistical and machine learning literature. 

Early research for the contextual bandit problem includes \cite{Wang/Kulkarni/Poor:2005} and \cite{Pandey/Agarwal/Chakrabarti/Josifovski:2007} and makes additional assumptions about the player's knowledge of the distribution or relationship between arms. One of the first practical algorithms to be discussed in the context of horizon-unaware side-information bandits was Epoch-Greedy presented by \cite{Langford/Zhang:2007}. One of the most salient works in this space is that of \cite{Dudik/Hsu/Kale/Karampatziakis/Langford/Reyzin/Zhang:2011} which brings contextual learning to a practical light by producing an online learning algorithm with a running time in $polylog(N)$ and regret that is additive in feedback delay. Additionally, the work of \cite*{Chu/Li/Reyzin/Schapire:2011} produces an analysis of an intuitive linear model-type upper confidence bound solution called LinUCB \cite{Dani/Kakade/Sham/Hayes:2007,Rusmevichientong/Tsitsiklis:2010,Abbasi-Yadkori/Szepesvari:2011} derived from the UCB solutions for non-contextual bandits which provides good real world performance.

Importantly, Exp4 \cite{Auer/Cesa-Bianchi/Freund/Schapire:2002} makes no statistical assumptions about the state of the world or arms and therefore can be applied to the contextual problem, however, the majority research thus far derived from Exp-type algorithms has been focused on the adversarial problem discussed prior. One exception is the application of Exp4.P to the strictly contextual problem found in \cite{Beygelzimer/Langford/Li/Reyzin/Schapire:2010}. Unfortunately, while achieving many of the goals of contextual bandits (and adversarial bandits) Exp4's running time scales linearly with the number of possible ``experts'' or contextual options available to it. This makes it inappropriate for continuous or highly optional contexts.

An in-depth mathematical survey of algorithms for the contextual multi-armed bandits is presented in \cite{Zhou:2015} from the perspective of the existing bandits literature.


Returning briefly to the slot machine example, contextual bandits model the situation where the machines have properties (arm-context) we believe may effect their payoff: perhaps some machines are red, some machines are blue (categorical context); perhaps machines closer to the front casino seem to pay better (linear, continuous context). This could also represent the situation where payoffs vary by day of week, time of day or another (world-context): perhaps slot machines in general are set by the casino to pay more on weekdays than on weekends, in effort to increase the number of plays during the week.

\subsubsection{LinUCB}
\label{policy:linUCB}
LinUCB \cite{Li/Chu/Langford/Schapire:2010} is a strong, intuitive polynomial time approach to the contextual bandits problem. Largely, LinUCB builds on the upper confidence bound work of the non-contextual bandits solution by synthesizing concepts captured by the associative reinforcement learning algorithm LinRel \cite{Auer:2003}. LinUCB introduces a feature vector to the UCB estimate which is maintained with technique very similar to a ridge regression.

In general form, LinUCB observes a set of $d$ features per arm (i) $x_{t,i}$ at each time step (t) and then selects an arm by a maximization of the regularized upper confidence bound estimate,

\begin{equation}
i = \argmax_i \theta^{'}_t  x_{t,i} + \alpha \sqrt{x_{t,i}^{'} A^{-1} x_{t,i}}
\end{equation}

where $\alpha$ is a positive regularization parameter and $\theta_t$ is the coefficient estimate for the arm's features. ($\theta_t = A^{-1}b$ where $A$ and $b$ are maintained via the ridge regression updating process after observing the reward\footnote{Recall that in the regression minimization problem, $\hat{\theta} = (X^{'} X)^{-1} X^{'} y$ and let $A = X^{'} X$ and $b=X^{'} y$ where $y$ is the observed reward})

LinUCB achieves regret in $\mbox{polylog}(H)$. Specifically, the regret bound shown by \cite*{Chu/Li/Reyzin/Schapire:2011} is $O(\sqrt{Td \ln^3(KH \ln(H)/\delta)})$ for $d$ dimensional feature vectors up to a probability of $1-\delta$. The algorithm's sensitivity to non-stationarity and feedback delay has not yet been investigated in depth though it may perform adequately on feedback delayed situations as the effect (or ``pull'') of each additional observation should decrease in increasing trials.  

A related algorithm discovered in the same year provides a model similar to LinUCB for the generalized linear model in \cite{Filippi/Cappe/Garivier/Szepesvari:2010} with finite time bounds dependent on the dimension of the parameter space. Related work was also provided by the linearly parameterized bandits paper of \cite{Rusmevichientong/Tsitsiklis:2010}.

\subsubsection{LinTS}
LinTS \cite{Burtini/Loeppky/Lawrence:2015a} provides a mechanism for varying linear bandits away from the UCB strategy. Recent research has shown Thompson sampling (TS) strategies to have empirical and theoretical performance that is competitive with the best known non-TS algorithms \cite{Chapelle/Li:2011, Munos:2014, Kaptein:2015}. 

In simple LinTS, a linear model is fitted with the traditional ordinary or penalized least squares procedures upon a design matrix made up of dummy variables for each arm, any interaction variables and any other modelling considerations desired by the statistician.  Estimates of the variance-covariance matrix and parameter coefficients are computed and used in a Bayesian fashion to produce an estimate of the current payout. For instance, if $\xi$ is the vector of covarying contextual variables and $A$ the list of arms, a model of the form,

\begin{equation}
Y_t = \alpha + \beta A + \gamma A \xi + \epsilon_t
\end{equation}

can be fitted and used to produce a normally distributed estimate of the reward at time $t+1$ by iterating through each arm and the contextual factors $\xi$ and computing both the mean payoff (for each arm $i$, $\hat{\alpha} + \hat{\beta} A_{i,t} + \hat{\gamma} A_{i,t} \xi_{i,t}$) and the standard error of the mean payoff by aggregating the values in the variance-covariance matrix according to the well-known sum of variances formula ($\sum_{i=1}^n \sum_{j=1}^n \mbox{Cov}(X_i, X_j)$).

\subsubsection{CoFineUCB} 

An interesting approach to the contextual bandits problem is to treat the exploratory contexts as a hierarchy. When this works, it could achieve logarithmic treatment of the features by treating them as a tree. Generalizing LinUCB, CoFineUCB  \cite{Yue/Hong/Guestrin:2012} approaches the estimation in a coarse-to-fine approach that allows increasing accuracy by drilling into a particular variable subspace. CoFineUCB extends LinUCB to fit a model strictly within a selected ``coarse'' subspace with a regularization parameter for the ``fine'' regression. The intuition provided is one of user's preferences -- if preferences can be embedded in a coarse-fine hierarchy (e.g., movies (coarse), action movies (fine); or ice cream (coarse), vanilla ice cream (fine)), then an initial model on the coarse levels can be supplemented by a stronger model on only those within the class to predict the fine levels.

In practice, CoFineUCB has been used in a recommender system context and shows good performance on experimental measures of regret when the coarse subspace accurately reduces the prediction variation for most users.

\subsubsection{Banditron and NeuralBandit}
The contextual bandits solutions explored so far require the effect of context be linear in the parameters within the interval being estimated. While some flexibility exists in terms of acceptable error rate and interval estimation, the linear regression techniques are all subject to similar constraints. Banditron \cite{Kakade/Shalev-Shwartz/Tewari:2008} and NeuralBandit \cite{Allesiardo/Feraud/Bouneffouf:2014} are recent algorithms for the non-linear contextual bandit which utilize the insights from the multi-layer perceptron \cite{Rosenblatt:1958}. At a high-level, these algorithms replace the (generally back-propagation based) updating process in the perceptron algorithm, with a partial information technique using only the bandit feedback. The specific update process differs in each algorithm.
 
As of the date of this work, neural network-based techniques lack much theoretical analysis and show significantly suboptimal regret in stationary and linear applications, however they are robust to both non-stationarity and non-linearity (and do not require a convex cost function whatsoever) where they show superior results.

\subsubsection{RandomizedUCB and ILOVETOCONBANDITS}
A UCB-based algorithm called RandomizedUCB is presented by \cite{Dudik/Hsu/Kale/Karampatziakis/Langford/Reyzin/Zhang:2011}. Unlike the algorithms presented thus far in this section, RandomizedUCB achieves the contextual lower bound of regret with regret of $O(\sqrt{HK \ln {HK/\delta}})$ with a probability of at least $1-\delta$ while maintaining a polylogarithmic computation time. RandomizedUCB requires calls to an optimization oracle on the order of $\tilde{O}(H^5)$ over the $H$ plays.

Advancing RandomizedUCB, \cite{Agarwal/Hsu/Kale/Langford/Li/Schapire:2014} produce what can be considered the state-of-the-art in contextual bandit algorithms with respect to regret and computational complexity with their algorithm \emph{Importance-weighted LOw-Variance Epoch-Timed Oracleized CONtextual BANDITS} algorithm (ILOVETOCONBANDITS). ILOVETOCONBANDITS produces an epoch-based coordinate-descent type approach to RandomizedUCB to reduce the number of calls to the optimization oracle to $\tilde{O}(\sqrt{KH/\ln{(N/\delta)}})$ with probability $1-\delta$.

Fundamentally, ILOVETOCONBANDITS exploits the structure in the policy space in order to treat it as an optimization problem. This is implicitly the technique taken in strategies like LinUCB (and the GLM or TS variants) and many other contextual bandit strategies. This alone is insufficient to produce an efficient algorithm, as it does not itself solve the exploration problem effectively. ILOVETOCONBANDITS is dependent on an estimate of a policy's reward calculated from the prior history of interaction-reward pairs using \emph{inverse propensity scoring} to produce importance-weighted -- that is, weighted on some randomization distribution $p_t$ --  estimates of the reward for each time $t$. The importance-weighting distribution $p_t$ is selected according to a set of updating rules which produce the low regret and low variance constraints necessary to reach optimality.

Algorithmically, ILOVETOCONBANDITS requires an epoch schedule which determines how frequently to recompute the optimization problem. At each iterate, an action is drawn in a style similar to Thompson sampling from an adjusted distribution of reward-weighted policies, and then upon reaching a new epoch, the distribution across all possible policies $Q$ must be updated to meet a bound on the empirical regret $\sum_\pi Q(\pi) b_\pi \leq 2K$ where $b$ is the scaled regret for the current policy $\pi$ and a similar bound on the variance. The details of the solution to this optimization problem with coordinate descent is shown in \cite{Agarwal/Hsu/Kale/Langford/Li/Schapire:2014}.

Both RandomizedUCB and ILOVETOCONBANDITS have a large burden in the details and complexity of implementation with no known public implementations available to date, but provide strong theoretical guarantees with both efficient computationally bounds and efficient regret bounds.

\subsection{Nonstationary Bandits for Ongoing Experimentation}

Non-stationary bandit problems is currently a very active research area. Slowly changing environments have been explored in depth in the Markov decision process literature \cite{Sutton:1990,Dearden:2000}. In their paper \cite{Garivier/Moulines:2008}, Garivier and Moulines prove an $O(\sqrt{n})$ lower-bound of regret for generalized non-stationary bandits. Staying with the medical trials context, we could imagine an ongoing study to determine which of a number of antibiotics to select from - when a clinician determines a particular class of antibiotics is appropriate for a patient, an ongoing experimental context may be used to select from the set of available products in a world where resistances, community bacterial loads and other factors may be evolving in an unmodelled or inherently non-stationary way.


\subsubsection{Discounted UCB(-T)}

Discounted UCB and Discounted UCB-Tuned \cite{Kocsis/Szepesvari:2006,Garivier/Moulines:2008} build on the work of UCB1 and UCB-Tuned for the original stochastic bandit problem, modifying the uncertainty padding estimate (the second term in the maximizing condition) and using a ``local empirical average'' instead of the traditional average considering older data in a discounted fashion. Effectively, discounted UCB creates an exponentially decayed version of UCB parameterized by some discount factor $\gamma \in (0,1)$.

In the same fashion as the original UCB, in Discounted UCB, at time $t$ we select the arm $i$ that maximizes the form $\bar{x}_{i,t} + c_{i,t}$ where $c_{i,t}$ is a measure that ``shifts'' the estimate upward (often selected as a variance-adjusted estimator of the ``exploratory value'' of arm $i$). In Discounted UCB, however, we parameterize both terms of that equation with a discount factor $\gamma$. We use an indicator function $\mathbbm{1}_{\sigma}$ defined as $1$ if the condition $\sigma$ is true and $0$ otherwise and a list $A_t$ that indicates the arm selected at time $t$. Specifically,

\begin{equation}
\bar{x}_{i,t}(\gamma) = \frac{1}{N_t(\gamma, i)} \sum_{s=1}^t \gamma^{t-s} x_s(i) \mathbbm{1}_{A_s=i}
\end{equation}

where $N_t(\gamma, i)=\sum_{s=1}^t \gamma^{t-s} \mathbbm{1}_{A_s=i}$ is the discounted average denominator and $x_s(j)$ is the payoff received from arm $i$ so far. This equation serves to capture the mean \emph{discounted} payoff estimate for arm $i$ at time $t$. And $c_{i,t}$ is,

\begin{equation}
c_{i,t}(\gamma) = 2B \sqrt{\frac{\xi \log \sum_{j=1}^{K} N_t(\gamma,j)}{N_t(\gamma,i)}}
\end{equation}

where $B$ is an upper-bound on the reward, as in the general formulation of UCB1 and $\xi$ is a parameter selected as $\frac{1}{2}$ in their paper, but with little further exploration.

\cite*{Garivier/Moulines:2008} shows that Discounted UCB achieves optimum non-stationary regret up to logarithmic factors, $\tilde{O}(\sqrt{n})$. By replacing the $c_{i,t}$ term with the tuned term from UCB-Tuned with an additional time discounting in the same $\gamma^{t-s}$ fashion, we get a variant of Discounted UCB, Discounted UCB-Tuned that is expected to have the same empirical improvements as in the non-discounted case.

\subsubsection{Sliding-Window UCB(-T)}

Sliding-Window UCB (SW-UCB) \cite{Garivier/Moulines:2008} is an extension of Discounted UCB to use a sliding window rather than a continuous discount factor. A sliding window can be modelled as a discount factor of $100\%$ for all data points older than some parameter $\tau$ representing the size of the window. To define the UCB functions for SW-UCB, \cite{Garivier/Moulines:2008} extends the same UCB1-type maximization process  $\bar{x}_{i,t} + c_{i,t}$ for a $\tau$ period window as,

\begin{equation}
\bar{x}_{i,t}(\tau)= \frac{1}{N_t(\tau)} \sum_{s=t-\tau+1}^t x_s(i)
\end{equation}

Where $N_t(\tau) = \min(t, \tau)$ is the total length of the set being summed over, eliminating the discounting consideration from above\footnote{In their paper, $N_t$ is erroneously provided as the same discounted version provided for discounted UCB. This cannot be correct, as $\gamma$ is no longer provided and the average would be incorrect.}. The padding or optimism function, $c_{i,t}$, then, is,

\begin{equation}
c_{i,t}(\tau) = B \sqrt{\frac{\xi \log(\min(t, \tau))}{N_t(\tau, i)}}
\end{equation}

Where $N_t(\tau, i)$ indicates the number of times arm $i$ was played in the window of length $\tau$. SW-UCB performs slightly better in their experimentation than the pure discounted approach and has the benefit of not requiring maintenance of data older than $\tau$ records. Both algorithms are strongly superior in regret to the Exp3 type algorithms and UCB1 with no non-stationarity modifications for the non-stationary problems tested.

\subsubsection{Weighted Least Squares}
Burtini et al. \cite{Burtini/Loeppky/Lawrence:2015a} provide a modification to the linear model Thompson sampler (LinTS) which replaces ordinary least squares with a weighted least squares (WLS). Rather than using WLS to correct for strictly heteroskedastic measurements, the algorithm aims to correct for the reduced \emph{value} of the information for predicting at a given time. Weights are generated from a functional form dependent only on time\footnote{The algorithm itself is indifferent between collecting time as in calendar time or time as in number of trials, and indeed, the particular application of the algorithm will determine which is more appropriate. In many cases, especially when the frequency of trials is low and nonstationarity is believed to come from some exogenous process, number of minutes/hours/days will be a more appropriate decay dimension than number of trials.} for each observation prior to each fit of the linear model. Empirical results are presented which show that linear decay (weights as a function $w(t) = \frac{1}{ct}$ where $c$ is an parameter) and exponential decay (weights as a function $w(t) = \frac{1}{c^t}$) both show promise in general unknown environments.

\subsubsection{Adapt-EvE}
Hartland et al. \cite*{Hartland/Gelly/Baskiotis/Teytaud/Sebag:2006} present an extension to the UCB-Tuned algorithm \cite{Auer/Cesa-Bianchi/Fischer:2002} to deal with abrupt changes in the distribution associated with each arm. Adapt-EvE is considered a \emph{meta-bandit} algorithm in that it uses a bandit algorithm at a higher level of abstraction to determine which bandit algorithm parameterization to use at each time. In particular, Adapt-EvE works by running the UCB-Tuned policy until a change-point in the underlying distribution is detected using one of many changepoint detection algorithms (in their paper they use the Page-Hinckley test \cite{Page:1954,Hinkley:1969,Basseville:1988} with ``discounted inertia'' to only trigger in the change-point case, not the drifting case\footnote{The rationale presented in \cite{Hartland/Gelly/Baskiotis/Teytaud/Sebag:2006} for discounting the change-point statistic is that UCB-Tuned is capable of handling slowly drifting reward distributions within itself.}). Upon detecting a changepoint, a meta-bandit is initialized with two arms: one, which continues using the trained version of UCB-Tuned, and the other which resets all parameters and instantiates a new instance of Adapt-EvE. Training continues at the meta-bandit level (learning whether to continue using the trained data or learn again) and at the selected sub-level.

Since Adapt-EvE, some advances in changepoint detection have arisen such as the improved Page-Hinckley test proposed in \cite{Ikonomovska:2012} and further research on the adaptive bandit problem using Bayesian changepoint detectors has been conducted with success by \cite{Mellor/Shapiro:2013}. We have conducted an experiment across a diverse set of binomial and normally-distributed stochastic arms finding a small but consistent benefit to the improved Page-Hinckley test in Adapt-EvE. The ideas of meta-bandits and changepoint detection bandits presented in Adapt-EvE give rise to a general technique for handling nonstationarity where parameters such as bandit algorithm\footnote{In particular, since the 2006 paper introducing Adapt-EvE, KL-UCB \cite{Maillard/Munos/Stoltz:2011}, Bayes-UCB \cite{Kaufmann/Cappe/Garivier:2012} and Thompson sampling, as well as the entire space of contextual methods like LinUCB \cite{Li/Chu/Langford/Schapire:2010} have arisen and seem likely to be fruitful in handling the same types of uncertainty Adapt-EvE did.} and detector can be substituted for the problem at hand.

\subsubsection{Exp3.R}
Combining the core idea of changepoint detection with the exponential decay policy of Exp3 appropriate for adversarial problems, \cite{Allesiardo/Feraud:2015} produce a bandit policy called \emph{Exp3 with Resets} that is capable of handling switching or changepoint nonstationary adversarial bandit problems.

After an exposition on both Adapt-EvE and Exp3, Exp3.R is easy to understand: after a parametric number of purely egalitarian exploration-observations (called $\gamma$-observations from the parameter $\gamma$ controlling egalitarianism in Exp3) is observed within an epoch (called an interval), a new epoch is started and a changepoint detection test is run and if it detects a change, all Exp3 data is reset. In this algorithm, rather than using Page-Hinckley or a variant, the changepoint detector integrates the concept of \emph{optimism} by only detecting if the empirical mean of any arm in this interval \emph{exceeds} the empirical mean of the believed best-arm by a parametrized amount. This optimistic changepoint detector uses the same principle as optimism in Thompson sampling, recognizing that a downward change in non-optimal arms is not relevant to the best-arm detection problem.

Unlike Adapt-EvE, Exp3.R's presentation includes a theoretical analysis, with a regret bound of $O(N \sqrt{H \log H})$ where $N$ is the number of changepoints in which the best arm changes. Furthermore, in empirical work, Exp3.R appears to outperform Adapt-EvE in switching nonstationary, but otherwise purely stochastic experiments.

\subsubsection{Kalman Filtered Bandit}

Kalman filtered bandits \cite{Berg:2010,Granmo/Berg:2010,Granmo/Glimsdal:2011} have been investigated in which the estimate of the mean payout of an arm is maintained by a recursive sibling Kalman filter parameterized by two \emph{a priori noise} estimates $\sigma^2_{ob}$ for \emph{observation noise} or measurement error and $\sigma^2_{tr}$ for \emph{transition noise} (the non-stationarity error). Results are somewhat sensitive to these noise estimates. At each time step $t$, an estimate of the mean and variance for the arm played (with reward received $x_{i,t}$) is updated,

\begin{equation}
\mu_{i,t} = \frac{(\sigma_{i,t-1}^2 + \sigma_{tr}^2) \cdot x_{i,t} + \sigma_{ob}^2 \cdot \mu_{i,t-1}}{\sigma_{i,t-1}^2 + \sigma_{tr}^2 + \sigma_{ob}^2},
\end{equation}
\begin{equation}
\sigma^2_{i,t} = \frac{(\sigma^2_{i,t-1} + \sigma_{tr}^2) \cdot \sigma_{ob}^2}{\sigma_{i,t-1}^2 + \sigma_{tr}^2 + \sigma_{ob}^2}.
\end{equation}

The non-played arms all have $\sigma_{tr}^2$ added to their variance estimate for each time step, indicating how their uncertainty increases as time progresses. These equations and the general form of this model arise from the well-studied Kalman filter. The numerous published extensions to the Kalman filter for varying confounding factors can likely be applied in this space.

This approach performs very well in drifting and change-point cases, however is outperformed by Adapt-EvE in the well-defined change-point case. The resilience to form of non-stationary make this a valuable approach in the event the parameters can be well predicted. This has not been explored within a contextual context, with Thompson sampling or probability matching techniques or with an optimistic approach. 


\subsection{Infinite- and Continuum-Armed Bandits for Continuous Spaces}
Expanding on the traditional model is a variant which treats the number of arms as an infinite or continuous range with some functional form defining the relationship \emph{or} a sufficient mechanism for discretizing the infinite space. This variant allows for substantial variation in problem difficulty by varying how much the agent knows about the arm relationship.

As an example of the infinite-armed bandit case, consider the case of picking a color for a purchase button to optimize clicks. Each user that views the purchase button is (possibly) influenced by its color, and color is a (theoretically) continuous function. As it would be impossible to sample all the colors, in the infinite-armed case, for the analysis of an infinite-armed bandit to be tractable, there must exist an underlying well-behaved function defining the relationship between arms (colors) and the payoff function (clicks). 


Recent work has studied such variants of the infinite armed problem as high-dimensional spaces \cite{Tyagi/Stich/Gartner:2014,Tyagi/Gartner:2013}, non-smooth spaces \cite{Combes/Proutiere:2014} and multiple-objectives \cite{vanMoffaert/vanVaerenbergh/Vrancx/Nowe:2014}, although work on theoretical analysis of the existing algorithms is still ongoing. An approach to define a computationally and mathematically feasible regret in a generalized infinite-armed case is presented in \cite*{Carpentier/Valko:2015}.

\subsubsection{Bandit Algorithm for Smooth Trees (BAST)}
\label{BASTsection}

Coquelin and Munos \cite{Coquelin/Munos:2007} present an analysis of the popular UCT (upper confidence bound for trees) algorithm  which combines Monte Carlo tree-based techniques from artificial intelligence\footnote{A good survey of related algorithms for tree search is given by Browne et al. \cite{Browne/Powley/Whitehouse/Lucas/Cowling/Rohlfshagen/Tavener/Perez/Samothrakis/Colton:2012}.} with the UCB1 algorithm discussed prior. UCT is popular in planning problems for game-playing artificial intelligence, but not itself appropriate for infinite-armed bandit problems as it applies in a space where the number of potential decisions is extremely large\footnote{Many games are well modelled as a tree structure where each decision reveals a new set of decisions until an end state (win or lose) is reached. These trees can grow very rapidly, but are not generally continuous. A number of techniques are proposed for dealing with them \cite{Pearl:1984} and these techniques frequently overlap with some techniques proposed for infinite- or continuum-armed bandits.}, not infinite.

The Bandit Algorithm for Smooth Trees introduces a mechanism related to continuity in to the tree approach, choosing to, for example, represent a continuous space as a tree with repeated branches dividing that space. The fundamental assumption is that related leaf nodes in the tree can be expected to have similar values in the payoff space. Coquelin and Munos \cite{Coquelin/Munos:2007} represent this assumption rigorously requiring that for any level in the tree, there exists a value $\delta_d > 0$ called the smoothness coefficient such that for at least one (optimal) node $i$ in that level $d$ the gap between the optimal leaf node ($\mu^*$) and all other leaf nodes is bounded by $\delta_d$. Formally, \begin{equation}\mu^* - \mu_j \leq \delta_d \quad \forall j \in \mbox{leaf(i)}\end{equation}

Assumptions of this style are the core tool with which infinite-armed bandits are generally represented. The tree is then generally assumed to take a coarse-to-fine hierarchical representation of the space down to some maximum tree height.

BAST performs selection at each level of the tree using a variant of the UCB algorithms which takes in to consideration the estimates of nearby nodes. Specifically, for a given smoothness coefficient $\delta_d$ for each level on the tree, a selection mechanism for any non-leaf node is given as maximizing

\begin{equation}
	B_{i, n_i} = \min \left\{ (\max B_{j, n_j}), X_{i, n_i} + \delta_d + c_{n_i} \right\}
\end{equation}

And for any leaf node as the simple UCB criteria, 

\begin{equation}
	B_{i, n_i} = X_{i, n_i} + c_{n_i}
\end{equation}

Where $c_{n_i}$ takes the role of the padding function with $n_i$ the number of times node $i$ has been visited from UCB and is defined as 
\begin{equation}
	c_n = \sqrt{\frac{2\log(Nn(n+1)\beta^{-1})}{2n}}
\end{equation}

A technique described in \cite{Coulom:2006} and \cite{Gelly/Wang/Munos/Teytaud:2006} is also explored in \cite{Coquelin/Munos:2007} which allows the online production of the tree with or without the assumption of a maximum tree height. In the iterative growing variant of BAST, the algorithm starts with only the root node, then at each iterate, selects a leaf node via its selection mechanism described above and expands the node in to two new child nodes, immediately choosing to play each child node once. This iterative technique still requires $O(n)$ memory to maintain the tree, but results in only optimal branches being explored in depth, a desirable property.

\subsubsection{Hierarchical Optimistic Optimization (HOO)}
Similar to BAST, Hierarchical Optimistic Optimization \cite{Kleinberg/Slivkins/Upfal:2008,Bubeck/Munos/Stoltz/Szepesvari:2011,Bubeck/Munos/Stoltz/Szepesvari:2008} attempts to build an estimate of the functional form $f$ by treating the problem as a hierarchical (coarse-to-fine) tree, with a particular focus on only maintaining a high-precision estimate of $f$ near its maxima. HOO builds and maintains a binary tree where each level is an increasingly precise subspace of the total space of arms, $X$. Each node in the tree tracks its interval range (subspace), how many times the node has been traversed and the empirical estimate of the reward, which it uses to compute an optimistic upper-bound estimate, $B$, for this leaf's reward in a similar fashion to the UCB algorithms. At each time step, the algorithm traverses the tree, picking the highest $B$ node at each junction until it reaches a leaf. At a leaf, it splits the node and creates a new point which is evaluated and the upper-bound estimate is updated up the tree accordingly.


This makes an assumption about the shape of $f$, but not one as strong as the BAST algorithm did. Rather than requiring continuity in a strongly defined sense as in the $\delta_d$ existence assumption before, HOO requires only that a dissimilarity function exists which puts a lower bound on the mean-payoff function over the arbitrary space the arms exist in.

In HOO, the selection strategy for each node requires a measure $\mu_{d,i}$ which represents the empirical mean of the payoff from each time the node been traversed and $N_{d,i}$, the number of times that node has been traversed. We use $d$ to denote depth in the tree, as in the BAST exposition, and $i$ to denote the specific node. Following again in the UCB strategy, the corresponding upper confidence bound criterion can be given as \begin{equation}
\mu_{d,i} + \sqrt{\frac{2\ln n}{N_{d,i}}} + \upsilon_1 \rho^d
\end{equation} where $0 < \rho < 1$ and $\upsilon_1>0$ are parameters selected by the implementer \cite{Bubeck/Munos/Stoltz/Szepesvari:2008}.

\subsection{Bandits with Multiple Plays}
A novel type of problems is bandits with multiple plays. In this environment, $m$ arms are selected at each time $t$. This is related to special class of the feedback delay problem considered earlier, where the delay is fixed and batched in to chunks equivalent to $m$ \emph{distinct} plays, but regret must be considered differently, as a certain class of results is impossible to achieve under traditional definitions of regret. Applications of this variant usually arise in recommender systems and online advertising environments, where multiple distinct options must be selected simultaneously (for instance, displaying multiple advertisements on a single page or presenting multiple product recommendations). Another application of bandits with multiple plays involves sensors or testing equipment, where a finite (and relatively scarce) set of equipment must be used to find an unknown object or value (e.g., in robotics, using $m$ cameras to find a hidden object) in an environment.

\subsubsection{Exp3.M}


Uchiya et al. \cite{Uchiya/Nakamura/Kudo:2010} present Exp3.M as the best known algorithm for applying an adversarial approach to bandits in a multiple-plays environment. The idea of applying Exp3 in multiple plays has been explored by other authors, but generally requires an expensive computation as the Exp3 ``arms'' are expanded to be the number of potential action-sets, which is exponential in the size of the original action space, $K$, and the size of $m$.  Exp3.M remains in the original action space (size $K$) and presents a computationally efficient algorithm. Like Exp3, Exp3.M involves computing a weighted sample, updated exponentially.

The ideas of multiple play adversarial bandits are further developed with \cite{Bubeck/Cesa-Bianchi:2012} presenting a generalization for a form of the problem they call combinatorial bandits. \cite{Louedec/Chevalier/Mothe/Garivier/Gerchinovitz:2015} address some practical considerations of Exp3.M resulting in an implementation which runs in $O(K \log K)$ time.  Other Exp3-like algorithms exist for this problem (e.g., adaptions of the RBA or IBA algorithms \cite{Radlinski/Kleinberg/Joachims:2008} which treat the problem as a series of independent problems) but \cite{Uchiya/Nakamura/Kudo:2010} shows Exp3.M to converge to the optimal selection faster than other algorithms while remaining highly-computable and appropriate for general application without giving up the core theoretical guarantees.


\subsubsection{Thompson Sampling and MP-TS}
For the stochastic problem, \cite{Komiyama/Honda/Nakagawa:2015} present an extension of Thompson sampling called MP-TS (multiple play Thompson sampling) extended to the multiple plays case, applied in example for the online advertising environment where multiple distinct advertisements are to be shown on a single page. Their extension is simply traditional Thompson sampling where the top $m$ arms are taken from the sampling process at each iterate rather than the top $1$. They go on to prove that this is asymptotically optimal for a definition of regret adjusted to the multiple plays space and then conduct empirical tests which show this method to be superior to both Exp3.M and their own similar extension to the KL-UCB algorithm referred to as MP-KL-UCB.

Interestingly, they extend their analysis one step further showing an empirical improvement to a modified algorithm called IMP-TS (improved multiple play Thompson sampling) where in a multiple play iterate requiring $m$ selections, $m-1$ are taken from the highest empirical means and the $m$th arm is selected with traditional single play Thompson sampling. This technique slightly outperforms the MP-TS technique in their experiments while still maintaining the asymptotic bound. In particular, they show that when the number of rounds is small (on the order of thousands), the gains from IMP-TS are large.

While this is a very recently discovered algorithm, the trend towards Thompson sampling techniques in multi-armed bandits and its ability to achieve provably optimal regret with minimum computation and implementation cost is a strong benefit for MP-TS and IMP-TS for this variant of the problem.




\section{State of the Theory and Future Work}
While regret is a deeply studied consideration in the field of multi-armed bandits, there are a number of algorithms with currently unknown regret properties, as well as bounds that are not expected to be as tight as they could be. Regret for simple stochastic bandits has been known since the seminal \cite{Lai/Robbins:1985} paper to be bounded by $O(\log H)$, but this bound does not hold for many of the generalizations and variations of the bandit problem. \cite{Cesa-Bianchi/Dekel/Shamir:2013} provide an enumeration from the perspective of adversarial generalizations of the bandit problem of regret lower- and best upper-bounds (as well as advancing some of the lower bounds) which shows that for oblivious adversaries, the lower- and best upper-bounds of regret are in $O(\sqrt{H})$, for bandits with switching costs, the lower- and best upper-bound are in $O(H^{2/3})$ and for the worst case of an adaptive adversary, the lower- and upper-bounds are linear in $H$. Nonstationary bandits under simple assumptions have a theoretical regret lower bound of $O(\sqrt{H})$ \cite{Garivier/Moulines:2008}. When $\delta$ is present in the bound, it is given as a high-probability bound with probability at least $1-\delta$.

\begin{sidewaystable}[h]
\renewcommand{\footnotesize}{\scriptsize}

\centering
\caption{Known Regret Bounds}
\label{my-label}
\resizebox{0.9\columnwidth}{!}{\begin{tabular}{|l|c|c|c|c|c|}
\hline
Algorithm & Environment & Asymptotic Bound & Finite Bound & Notes \\
\hline
$\epsilon$-greedy (adaptive $\epsilon$) \cite*{Auer/Cesa-Bianchi/Fischer:2002} 	& K-armed, stochastic & $O(\log H)$ & $\sum_t (\frac{c}{d^2 t} + o(\frac{1}{t}))$ \cite*{Auer/Cesa-Bianchi/Fischer:2002} & \footnote{For the adaptive variant $\epsilon_n$-greedy given by \cite*{Auer/Cesa-Bianchi/Fischer:2002}. The bound given is the sum of instantaneous regrets (that is, proportional to the probability of drawing a suboptimal arm at time $t$) when $d$ is the lower bound on the gap between the best and second best expectations. The bound given applies only for ``large enough'' values of $c$ (e.g., $c>5$). The more general bound replaces the $o(\frac{1}{t})$ term with the fully determined term $2\left(    \frac{c}{d^2} \log\frac{(t-1)d^2 \sqrt{e}}{cK}   \right)
\left(   \frac{cK}{(t-1)d^2\sqrt{e}}    \right)^{c/(5d^2)}
+ \frac{4e}{d^2} \left(  \frac{cK}{(t-1)d^2\sqrt{e}} \right)^{c/2}$. This bound only applies for $t \geq \frac{cK}{d}$.}  \\

	
	
UCB1 \cite{Agrawal:1995} & K-armed, stochastic & $O(\log H)$ \cite*{Auer/Cesa-Bianchi/Fischer:2002} &   
$8 \cdot \left[{\sum_{i:\mu_i < \mu^*} \left(\frac{\log t}{\Delta_i}\right)}\right] + \left(1 + \frac{\pi^2}{3}\right)\left(\sum_{i=1}^K \Delta_i\right)$ \cite*{Auer/Cesa-Bianchi/Fischer:2002} 	 & \\
UCB2 \cite{Auer/Cesa-Bianchi/Fischer:2002} & K-armed, stochastic & $O(\log H)$ &

$\sum_{i: \mu_i < \mu^*} \left(\frac{(1+\alpha)(1+4\alpha)\log(2e\Delta_i^2t)}{2\Delta_i} + \frac{c_\alpha}{\Delta_i}\right)$ \cite*{Auer/Cesa-Bianchi/Fischer:2002} 	
&\\
UCB-Tuned \cite{Auer/Cesa-Bianchi/Fischer:2002} & K-armed, stochastic & -- & -- & \footnote{Empirical tests show that UCB-Tuned outperforms the earlier UCB variants, but no regret bound is proven.}\\
MOSS \cite{Audibert/Bubeck:2009}& K-armed, stochastic &$O(\log H)$ & $\min\left\{ \sqrt{tK}, \frac{K}{\Delta} \log \frac{t \Delta^2}{K} \right\}$ \cite{Audibert/Bubeck:2009}&\\
POKER \cite{Vermorel/Mohri:2005} & K-armed, stochastic & -- & --&\\
Bayes-UCB \cite{Kaufmann/Cappe/Garivier:2012} & K-armed, stochastic &$O(\log H)$& $\frac{1+\epsilon}{d(\mu_j, \mu^*)} \log(t) + o(\log(t))$ \cite{Kaufmann/Cappe/Garivier:2012} & \footnote{$d$ is the Kullback-Leibler divergence as defined in Equation \ref{KLdiv} and $\epsilon>0$.} \\
KL-UCB \cite{Garivier/Cappe:2011} & K-armed, stochastic &$O(\log H)$& &\\
Thompson sampling  (TS) \cite{Thompson:1933}        &  K-armed, stochastic                    & $O(\log H)$ & $O\left((\sum_{a=2}^{K} \frac{1}{\Delta_a^2})^2 \log t\right)$ \cite{Agrawal/Goyal:2012,Agrawal/Goyal:2013} &	\footnote{\cite{Agrawal/Goyal:2012} prove elegant bounds for the 2-armed bandit as well as a distinct variant of the finite-time bound with a better dependence on $t$. A bound is given with all constants (rather than Big Oh notation) in the same paper. The optimality of Thompson sampling is explored in depth by authors such as \cite{Honda/Takemura:2013}, \cite{Osband/Russo/vanRoy:2013} and \cite{Kaufmann/Korda/Munos:2012}. A different finite-time bound is given by \cite{Kaufmann/Korda/Munos:2012} which utilizes a problem-dependent constant and produces a similarly asymptotically optimal equation.}     \\
Optimistic TS \cite{Chapelle/Li:2011}         &      K-armed, stochastic                 	& -- & --& \footnote{Optimism as a bias in Thompson sampling performs better in empirical studies.}      \\
Bootstrap TS \cite{Eckles/Kaptein:2014}        &       K-armed, stochastic                	& -- & --&     \\
Simple (Optimistic) Sampler \cite{Burtini/Loeppky/Lawrence:2015b}        &    K-armed, stochastic & --                    & -- &	  \footnote{Nonparametric method, requiring no understanding of the underlying distribution.}      \\
BESA \cite{Baransi/Maillard/Mannor:2014}        &       K-armed, stochastic                	& $O(\log H)$ \cite{Baransi/Maillard/Mannor:2014} & $O(\log H)$ \cite{Baransi/Maillard/Mannor:2014} &  \footnote{Ibid.}    \\

\hline
Exp3 \cite*{Auer/Cesa-Bianchi/Freund/Schapire:2002}  & K-armed, adversarial & $\tilde{O}(\sqrt{KH})$ \cite*{Auer/Cesa-Bianchi/Freund/Schapire:2002} & -- & \footnote{A similar algorithm, called Exp3.P is given in which a similar bound holds in high-probability.} \\
Exp4 \cite{Auer/Cesa-Bianchi/Freund/Schapire:2002, Beygelzimer/Langford/Li/Reyzin/Schapire:2010} & K-armed, adversarial, contextual & $\tilde{O}(\sqrt{KH \log N})$ \cite*{Beygelzimer/Langford/Li/Reyzin/Schapire:2010} & -- & \footnote{The bound is proven for a variant of the algorithm called Exp4.P \cite*{Beygelzimer/Langford/Li/Reyzin/Schapire:2010} and is said to not hold for the basic Exp4 implementation.  The contextual variant in this case is ``with expert advice.''} \\
SAO \cite*{Bubeck/Slivkins:2012} & K-armed, adversarial & $O(\mbox{polylog}(H))$, $O(\sqrt{H})$ \cite*{Bubeck/Slivkins:2012}  & -- &\footnote{The regret bounds presented are for the stochastic case and the adversarial case respectively.} \\
\hline
LinUCB \cite{Li/Chu/Langford/Schapire:2010} & K-armed, contextual & ${O}(\sqrt{dH \frac{\ln{KH\ln{H}}}{\delta}})$

\cite{Li/Chu/Langford/Schapire:2010} & See Note & \footnote{$d$ as given here is the dimension of the design matrix used to produce the linear model. \cite{Li/Chu/Langford/Schapire:2010} claim the bound given follows directly from the work of \cite{Auer/Cesa-Bianchi/Fischer:2002} which may suggest a finite-time bound in the same order.} \\
LinTS \cite{Burtini/Loeppky/Lawrence:2015a} & K-armed, contextual & -- & -- & \\
RandomizedUCB \cite{Dudik/Hsu/Kale/Karampatziakis/Langford/Reyzin/Zhang:2011} & K-armed, contextual & $O(\sqrt{HK \ln(N/\delta)})$ \cite{Dudik/Hsu/Kale/Karampatziakis/Langford/Reyzin/Zhang:2011} & -- & \\
Banditron\cite{Kakade/Shalev-Shwartz/Tewari:2008} & K-armed, contextual, nonlinear & $O(H^{2/3})$ \cite{Kakade/Shalev-Shwartz/Tewari:2008} & -- & \footnote{In certain environments this can be improved to $O(\sqrt{H})$.} \\
NeuralBandit \cite{Allesiardo/Feraud/Bouneffouf:2014} & K-armed, contextual, nonlinear &-- & -- & \footnote{NeuralBandit outperforms Banditron and LinUCB in empirical experiments by \cite{Allesiardo/Feraud/Bouneffouf:2014} while also providing robustness to nonstationarity and nonlinearity.} \\
ILOVETOCONBANDITS \cite{Agarwal/Hsu/Kale/Langford/Li/Schapire:2014} & K-armed, contextual &  $O(\sqrt{HK \ln(N/\delta)})$ & -- & \\
\hline
Discounted UCB\cite{Garivier/Moulines:2008} & K-armed, nonstationary & $O(\sqrt{H})$ & $O(\sqrt{t \Gamma} \log t)$ \cite{Garivier/Moulines:2008} & \footnote{Where $\Gamma$ is the number of breakpoints in the regret history. Further assumptions on the frequency of breakpoints allow tighter bounds \cite{Garivier/Moulines:2008}.} \\
SWUCB \cite{Garivier/Moulines:2008} & K-armed, nonstationary & $O(\sqrt{H})$ & $O(\sqrt{t \Gamma \log t})$ \cite{Garivier/Moulines:2008} & \\
Adapt-EvE \cite{Hartland/Gelly/Baskiotis/Teytaud/Sebag:2006} & K-armed, nonstationary & -- & -- & \footnote{This showed the best empirical result in the Pascal-2 Exploration vs. Exploitation challenge in 2006.} \\
Exp3.R \cite{Allesiardo/Feraud:2015} & K-armed, nonstationary, possibly adversarial & $O(N \sqrt{H \log H})$ & -- & \\
WLS LinTS \cite{Burtini/Loeppky/Lawrence:2015a} & K-armed, nonstationary, contextual & -- & -- & \\
\hline
UCT \cite{Kocsis/Szepesvari:2006} & $\infty$-armed, continuous & $O(e^{e^D})$ \cite{Coquelin/Munos:2007} & -- & \footnote{Where $D$ is the depth of the tree used.} \\
HOO \cite{Bubeck/Munos/Stoltz/Szepesvari:2008} & $\infty$-armed, metric space & $\tilde{O}(\sqrt{H})$ \cite{Bubeck/Munos/Stoltz/Szepesvari:2011}  & -- & \\
\hline
ComBand \cite{Cesa-Bianchi/Lugosi:2009} & K-armed, adversarial, multi-play & $O(m^\frac{2}{3} \sqrt{H K \log K})$ \cite{Uchiya/Nakamura/Kudo:2010}  & -- & \\
BOLOM \cite{Uchiya/Nakamura/Kudo:2010} & K-armed, adversarial, multi-play & $O(m K^\frac{2}{3} \sqrt{H \log H})$ & -- &\\
Exp3.M \cite{Uchiya/Nakamura/Kudo:2010} & K-armed, adversarial, multi-play & $O(\sqrt{mH K \log(K/m})$ \cite{Uchiya/Nakamura/Kudo:2010} & -- & \footnote{The lower-bound for this algorithm appears as a generalization of the Exp3 bound when $m=1$. Furthermore, the same holds for the lower-bound as noted in \cite{Uchiya/Nakamura/Kudo:2010}: $\Omega(((K-m)/K)^2 \sqrt{KH})$.}\\
\hline
MP-TS \cite{Komiyama/Honda/Nakagawa:2015} & K-armed, multi-play & $O(\log H)$ \cite{Komiyama/Honda/Nakagawa:2015} &$O(\log t)$ &\footnote{This bound holds in high-probability as well.} \\
IMP-TS \cite{Komiyama/Honda/Nakagawa:2015} & K-armed, multi-play &--  & --& \footnote{IMP-TS shows better performance in empirical studies than MP-TS.} \\

         \hline

\end{tabular}}
\end{sidewaystable}

\FloatBarrier

Future work may explore how the objective function of minimizing regret can be combined with a constraint to produce sufficiently strong estimates of the underlying parameters of interest -- e.g., in the medical trials context, minimizing regret subject to the constraint of producing (in high probability) a desired level of confidence in each estimator. Such a fully parameterized constrained bandit implementation could produce significant value in terms of reducing the costs of \emph{ex-ante} uncertainty in high cost or high risk experiments.

\bibliographystyle{acmtrans-ims}
\bibliography{biblio}

\end{document}